\documentclass{article}

    \PassOptionsToPackage{numbers, compress}{natbib}

    \usepackage[preprint]{neurips_2023}

\usepackage[utf8]{inputenc} 
\usepackage[T1]{fontenc}    
\usepackage{hyperref}       
\usepackage{url}            
\usepackage{booktabs}       
\usepackage{amsfonts}       
\usepackage{nicefrac}       
\usepackage{microtype}      
\usepackage{subcaption}
\usepackage{caption}
\usepackage{wrapfig}
\usepackage{graphics}
\usepackage[pdftex]{graphicx}
\usepackage{amsmath,amssymb,bm}
\usepackage{multirow}
\usepackage{multicol}
\usepackage[table]{xcolor}

\usepackage{algorithm} 
\usepackage{algorithmic}

\definecolor{Blue9}{rgb}{0.098,0.3,0.9}
\definecolor{DarkBlue}{rgb}{0,0.08,0.45}

\hypersetup{colorlinks=true}
\hypersetup{citecolor=Blue9}
\hypersetup{linkcolor=Blue9}

\title{Language Reward Modulation for Pretraining Reinforcement Learning}

\author{
\textbf{Ademi Adeniji}
\vspace{1.0ex}
\qquad
\textbf{Amber Xie}
\vspace{1.0ex}
\qquad
\textbf{Carmelo Sferrazza}
\vspace{1.0ex}
\qquad
\textbf{Younggyo Seo}
\vspace{1.0ex}
\qquad \\
\textbf{Stephen James }
\qquad \vspace{3.0ex}
\textbf{Pieter Abbeel}
\\ \vspace{3.0ex}
UC Berkeley
}

\begin{document}

\maketitle
\begin{abstract}
Using learned reward functions (LRFs) as a means to solve sparse-reward reinforcement learning (RL) tasks has yielded some steady progress in task-complexity through the years. In this work, we question whether today's LRFs are best-suited as a direct replacement for task rewards. Instead, we propose leveraging the capabilities of LRFs as a pretraining signal for RL. Concretely,  we propose \textbf{LA}nguage Reward \textbf{M}odulated \textbf{P}retraining (LAMP) which leverages the zero-shot capabilities of Vision-Language Models (VLMs) as a \textit{pretraining} utility for RL as opposed to a downstream task reward. LAMP uses a frozen, pretrained VLM to scalably generate noisy, albeit shaped exploration rewards by computing the contrastive alignment between a highly diverse collection of language instructions and the image observations of an agent in its pretraining environment. LAMP optimizes these rewards in conjunction with standard novelty-seeking exploration rewards with reinforcement learning to acquire a language-conditioned, pretrained policy. Our VLM pretraining approach, which is a departure from previous attempts to use LRFs, can warmstart sample-efficient learning on robot manipulation tasks in RLBench. Source code is available at \url{https://github.com/ademiadeniji/lamp}.
\end{abstract}

\section{Introduction}
\label{intro}
A longstanding challenge in reinforcement learning is specifying reward functions. Extensive domain knowledge and ad-hoc tuning are often required in order to manually design rewards that ``just work.'' However, such rewards can be highly uninterpretable and riddled with cryptic mathematical expressions and constants. Furthermore, hand-crafted reward functions are often over-engineered to the domain in which they were designed, failing to generalize to new agents and new environments \citep{goalmisgen, goalmisgen2}. As a result, a long history of foundational work in Inverse Reinforcement Learning (IRL) \citep{irl, irl2, helicopter, amp, ampreward} has produced an abundance of methods for learning rewards from demonstration data assumed to optimal under the desired reward function. However, learned reward functions are also notorious for noise and reward misspecification errors \citep{aisafety, boat} which can render them highly unreliable for learning robust policies with reinforcement learning. This is especially problematic in more complex task domains such as robotic manipulation, particularly when in-domain data for learning the reward function is limited.

\begin{figure*}[t!]
    \centering
    \includegraphics[width=0.99\textwidth]{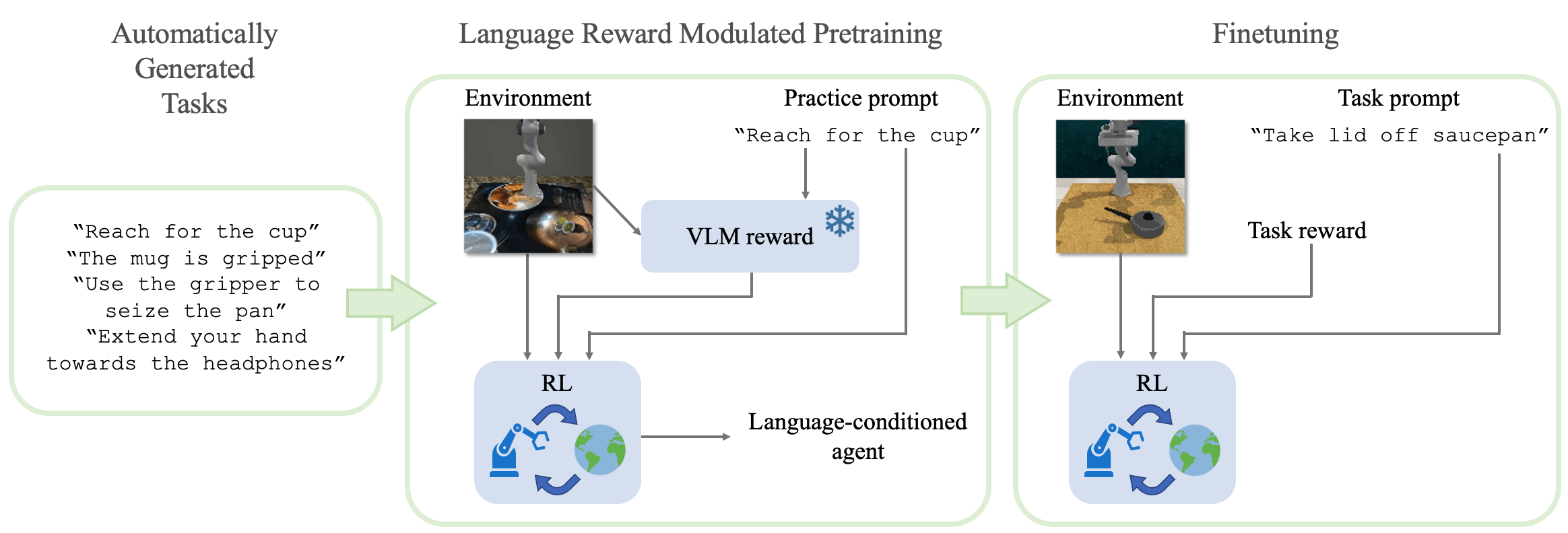}
    \caption{\textbf{LAMP Framework}. Given a diverse set of tasks generated by hand or by a LLM, we extract VLM rewards for language-conditioned RL pretraining. At finetuning time, we condition the agent on the new task language embedding and train on the task reward.}
    \label{fig:system}
\end{figure*}

While acknowledging that learned reward functions are subject to potential errors, we hypothesize that they may be effectively employed to facilitate exploration during the lower-stakes pretraining stage of training. During pretraining, we generally desire a scalable means of generating diverse behaviors to warmstart a broad range of possible downstream tasks. LRFs are an appealing means for supervising these behaviors since they do not rely on human-design and carry the potential to scale with dataset diversity. Despite this potential, obtaining LRFs that generalize to new domains is non-trivial \citep{goalmisgen}. Notably, however, large-pretrained models have shown impressive zero-shot generalization capabilities that enable them to be readily applied in unseen domains. Indeed, large pretrained VLMs have shown recent successes in reward specification for task learning by computing alignment scores between the image observations of an agent and a language input describing the desired task \cite{zest, zsrm}. While these methods adopt similar reward mispecification shortcomings to other LRFs, they come with the novel and relatively under-explored property of being a scalable means of generating many \textit{different} rewards by simply prompting with different language instructions. This property is particularly compatible with the assumptions of RL pretraining where we desire to learn general-purpose behaviors with high coverage of environment affordances, require minimal human supervision, and do not require pretrained behaviors to transfer zero-shot to fully solve downstream tasks. Instead of relying on noisy VLM LRFs to train task-specific experts, can we instead use them as a tool for pretraining a general-purpose agent?

In this work, we investigate how to use the flexibility of VLMs as a means of scalable reward generation to \textit{pretrain} an RL agent for accelerated downstream learning. We propose \textbf{LA}nguage Reward \textbf{M}odulated \textbf{P}retraning (LAMP), a method for pretraining diverse policies by optimizing VLM parameterized rewards. Our core insight is that instead of scripting rewards or relying solely on general unsupervised objectives to produce them, we can instead query a VLM with highly diverse language prompts and the visual observations of the agent to generate diverse, shaped pretraining rewards. We augment these rewards with intrinsic rewards from Plan2Explore \citep{plan2explore}, a novelty-seeking unsupervised RL algorithm, resulting in an objective that biases exploration towards semantically meaningful visual affordances. A simple language-conditioned, multitask reinforcement learning algorithm optimizes these rewards resulting in a language-conditioned policy that can be finetuned for accelerated downstream task learning. We demonstrate that by pretraining with VLM rewards in a visually complex environment with diverse objects, we can learn a general-purpose policy that more effectively reduces the sample-complexity of downstream RL. We train LAMP in a pretraining environment with realistic visual textures and challenging randomization and evaluate downstream performance on RLBench tasks. We also analyze the influence of various prompting techniques and frozen VLMs on the performance of the pretrained policy.

\section{Related Work}

\paragraph{Pretraining for RL} Following on the successes of pretraining in vision and language, a number of approaches have grown in interest for pretraining generalist RL agents in order to reduce sample-complexity on unseen tasks. Classical works in option learning \citep{options, optionsrl} and more recent approaches in skill discovery such as \citep{dads, vic, cds, diayn} look to pretrain skill policies that can be finetuned to downstream tasks. Exploration RL algorithms such as \citep{rnd, plan2explore, icm, apt} use unsupervised objectives to encourage policies to learn exploratory behaviors.  Works such as \citep{mvp, r3m, mvpreal} leverage pretrained vision encoders to accelerate RL from pixels. LAMP combines a large-pretrained VLM with exploration-based RL to guide exploration towards meaningful behaviors.

\paragraph{Inverse RL from human video}
Inverse reinforcement learning (IRL) \citep{irl_survey, mce, irl, irl2} proposes a number of approaches to address the challenges associated with learning reward functions from demonstrations. A number of more recent works focus on inverse RL from video datasets of human interaction, \citep{dvd, xirl, unsuppercept, tcn} which are often more readily available than in-domain demonstrations. These methods rely on perceptual metrics such as goal-image similarity and trajectory similarity to formulate rewards but require task-specific paired data. Other methods such as \citep{vip} make weaker assumptions on the task-specificity of the human video dataset and thus can leverage "in-the-wild" data and exhibit stronger domain generalization. LAMP similarly exploits "in-the-wild" video data via a frozen, pretrained VLM but focuses on leveraging language to flexibly modulate the VLM and generate diverse rewards.

\paragraph{VLMs as task rewards}
A number of works propose methods for extracting shaped task reward signals from large-scale pretrained LLMs or VLMs \citep{ellm, zest, zsrm}. Others such as \citep{successvqa} leverage pretrained VLMs as general-purpose success detectors which can provide sparse task rewards. VLM-based rewards can also be learned on bespoke datasets or in-domain data \citep{minedojo, concept2robot} to be used directly for downstream task learning. Instead of relying on VLMs to train task-specific experts, LAMP uses a VLM to control scalable RL pretraining and employs scripted task rewards to demonstrate reliable downstream finetuning.

\section{Background}
\label{sec:background}
\paragraph{Reinforcement learning} We consider the reinforcement learning (RL) framework where an agent receives an observation $o_{t}$ from an environment and chooses an action $a_{t}$ with a policy $\pi$ to interact with the environment.
Then the agent receives an extrinsic reward $r^{\texttt{e}}_{t}$ and a next observation $o_{t+1}$ from the environment.
The goal of RL is to train the policy to maximize the expected return defined as a cumulative sum of the reward with a discount factor $\gamma$, \textit{i.e.,} $\mathcal{R}_{t} = \sum_{k=0}^\infty \gamma^{k} r(o_{t+k}, a_{t+k})$.

\paragraph{Reinforcement learning with vision-language reward}
In sparse reward tasks, the extrinsic reward $r^{\tt{e}}$ becomes non-zero only when the task successfully terminates, making it difficult to learn policies that complete the target tasks.
To address this, recent approaches have proposed to use the vision-language alignment score from a large-scale vision-language model (VLM) as a reward \citep{minedojo,zest}.
Formally, let $\mathbf{x} := \{x_{1}, ..., x_{M}\}$ be a text that describes the task consisting of $M$ tokens, $F_{\phi}$ be a visual feature encoder, and $L_{\alpha}$ be a language encoder.
Given a sequence of transitions $\{o_{i}, a_{i}, r^{\tt{e}}_{i}, o_{i+1}\}_{i=1}^{N}$, the key idea is to use the distance between visual representations $F_{\phi}(o_{i})$ and text representations $L_{\alpha}(\mathbf{x})$ as an intrinsic reward, which is defined as $r^{\texttt{int}}_{i} = D\left(F_{\phi}(o_{i}), L_{\alpha}(\mathbf{x})\right)$, where $D$ can be an arbitrary distance metric such as cosine similarity or L2 distance.
This intuitively can be seen as representing the extent to which the current observation is close to achieving the task specified by the text.

\paragraph{R3M}
The vision encoders of video-language models have been successfully employed as semantic feature extractors that enable downstream learning on a variety of domains including standard prediction and classification tasks as well as, more recently, decision making and control \citep{videoclip}.
Notably, R3M, has lead to improvements in the data-efficiency of imitation learning in real-world robotics domains \cite{r3m}. R3M extracts semantic representations from the large-scale Ego4D dataset of language annotated egocentric human videos \citep{ego4d}. The language input is processed by $L_{\alpha}$, a pretrained DistilBERT transformer architecture \citep{distbert} that aggregates the embeddings of each word in the instruction and the images are encoded with R3M's pretrained ResNet-18 $F_{\phi}$. A video-language alignment loss encourages the image representations ${F}_{\phi}(\cdot)$ to capture visual semantics by extracting image features that aid in predicting the associated language annotations, which are embedded by $L_{\alpha}(\mathbf{x})$. In particular, R3M trains $G_{\theta}(F_{\phi}(o_1), F_{\phi}(o_i), L_{\alpha}(\mathbf{x}))$ to score whether the language $\mathbf{x}$ explains the behavior from image $o_1$ to image $o_i$.  The score function is trained simultaneously to the representations described above with a contrastive loss that encourages scores to increase over the course of the video and scores to be higher for correct pairings of video and language than incorrect ones. 

\begin{figure*}[t!]
    \centering
    \includegraphics[width=0.8\textwidth]{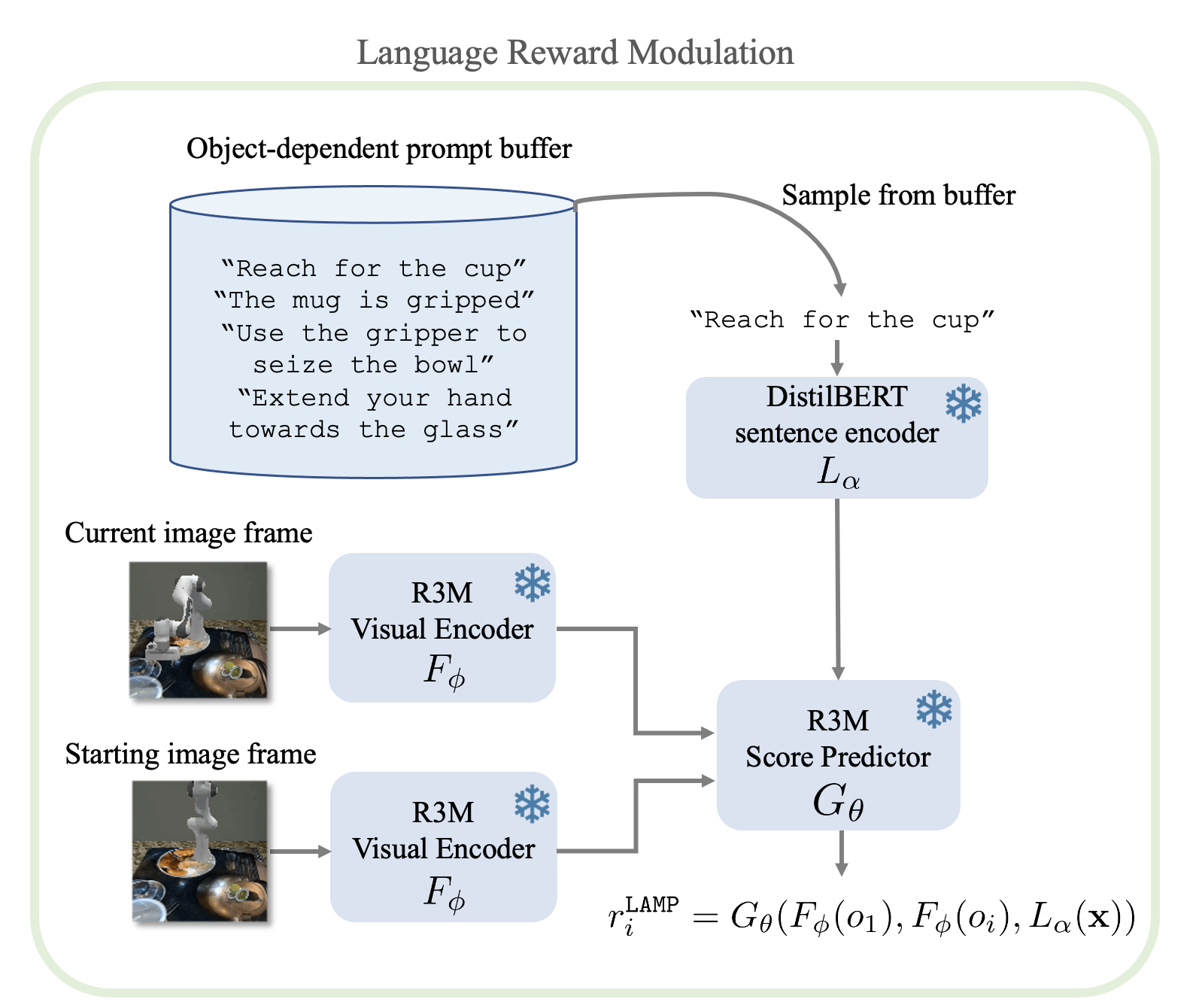}
    \caption{\textbf{LAMP Method}. We use R3M~\citep{r3m} for our VLM-based rewards. We query the R3M score predictor for pixel and language alignment, which is pretrained on the Ego4D dataset~\citep{ego4d}. The reward model is frozen.}
    \label{fig:method}
\end{figure*}

\section{Method}
We present \textbf{LA}nguage \textbf{R}eward \textbf{M}odulated \textbf{P}retraining (LAMP), a simple yet effective framework for pretraining reinforcement learning with intrinsic rewards modulated with language instructions.
LAMP consists of two stages: (i) a task-agnostic RL pretraining phase that trains policies to maximize the VLM-based rewards and (ii) a downstream task finetuning phase that adapts pre-trained policies to solve the target tasks by maximizing the task reward.
In this section, we first describe how we define our intrinsic reward (see Section~\ref{sec:method_reward}), how we pretrain policies (see Section~\ref{sec:method_behavior}), and how we adapt the policies to downstream tasks (see Section~\ref{sec:method_adaptation}).
We provide the overview and pseudocode of LAMP in Figure~\ref{fig:method} and Algorithm~\ref{alg:lamp}, respectively.

\subsection{Language Reward Modulation}
\label{sec:method_reward}

\begin{figure*}[t!]
    \centering
    \textbf{Video-Language Alignment Rewards}\par\medskip
    \includegraphics[width=0.3\linewidth]{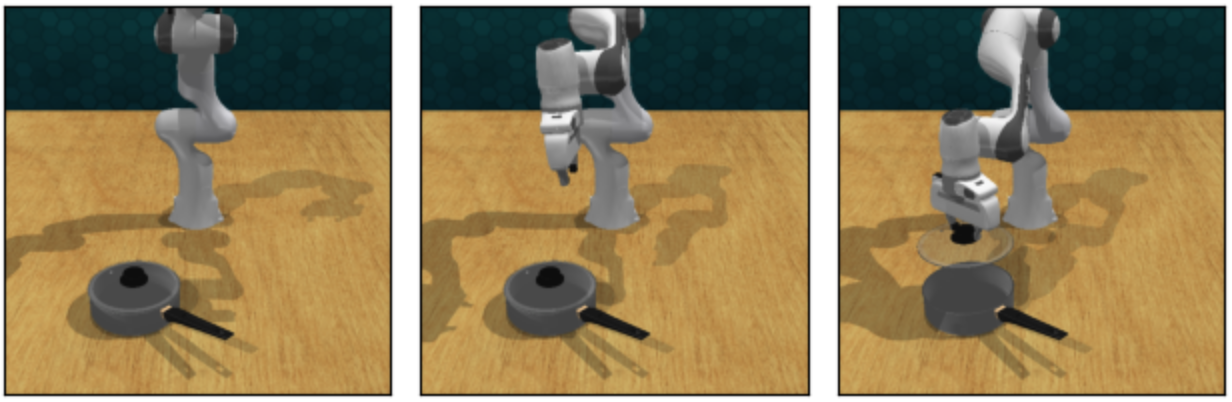}\hfill
    \includegraphics[width=0.3\linewidth]{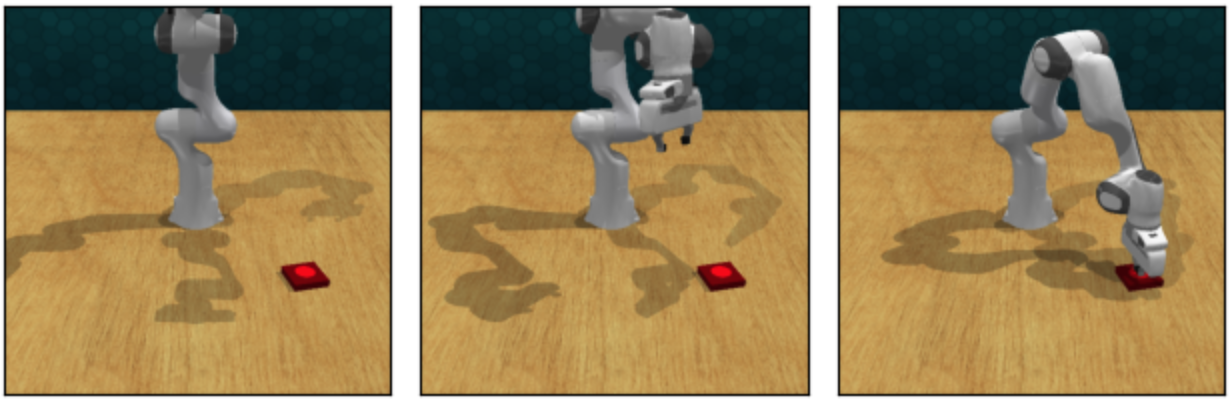}\hfill
    \includegraphics[width=0.3\linewidth]{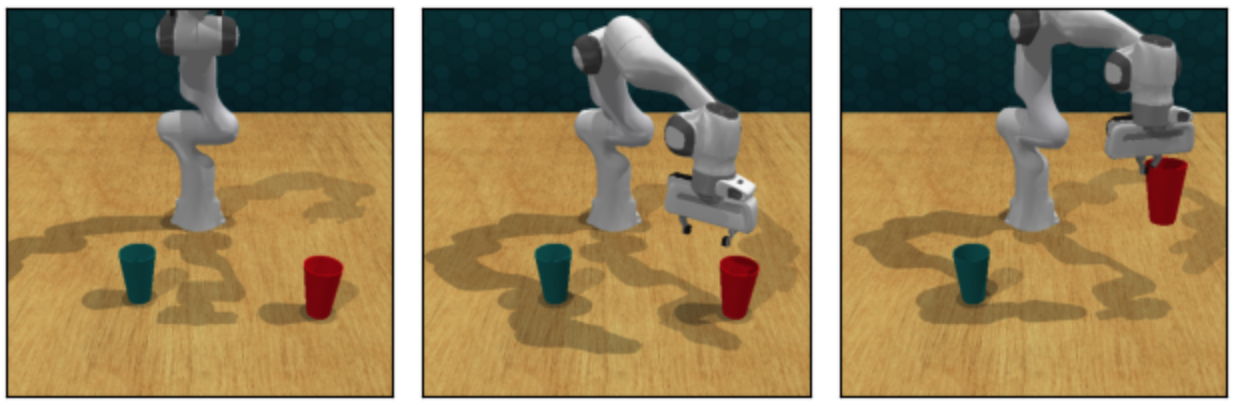}\par
    \vspace{10pt}
    \includegraphics[width=0.33\linewidth]{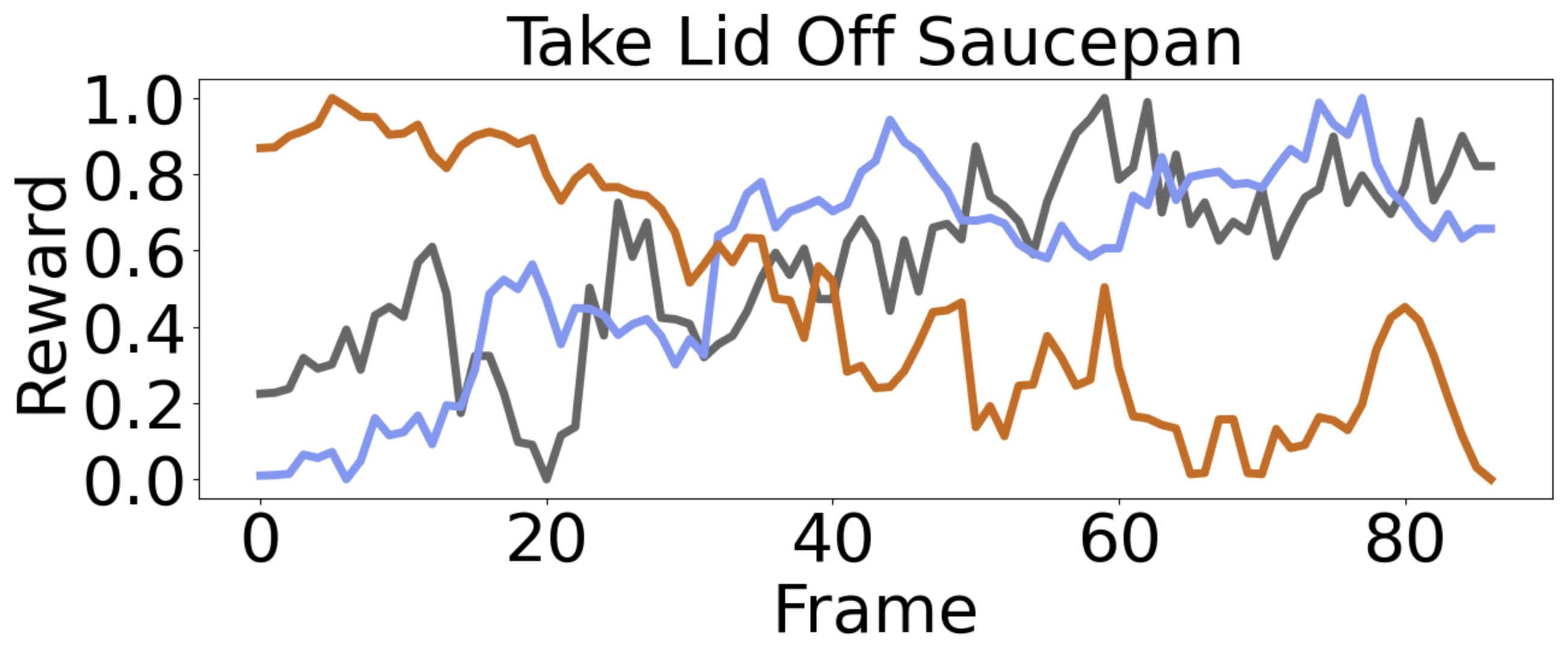}\hfill
    \includegraphics[width=0.33\linewidth]{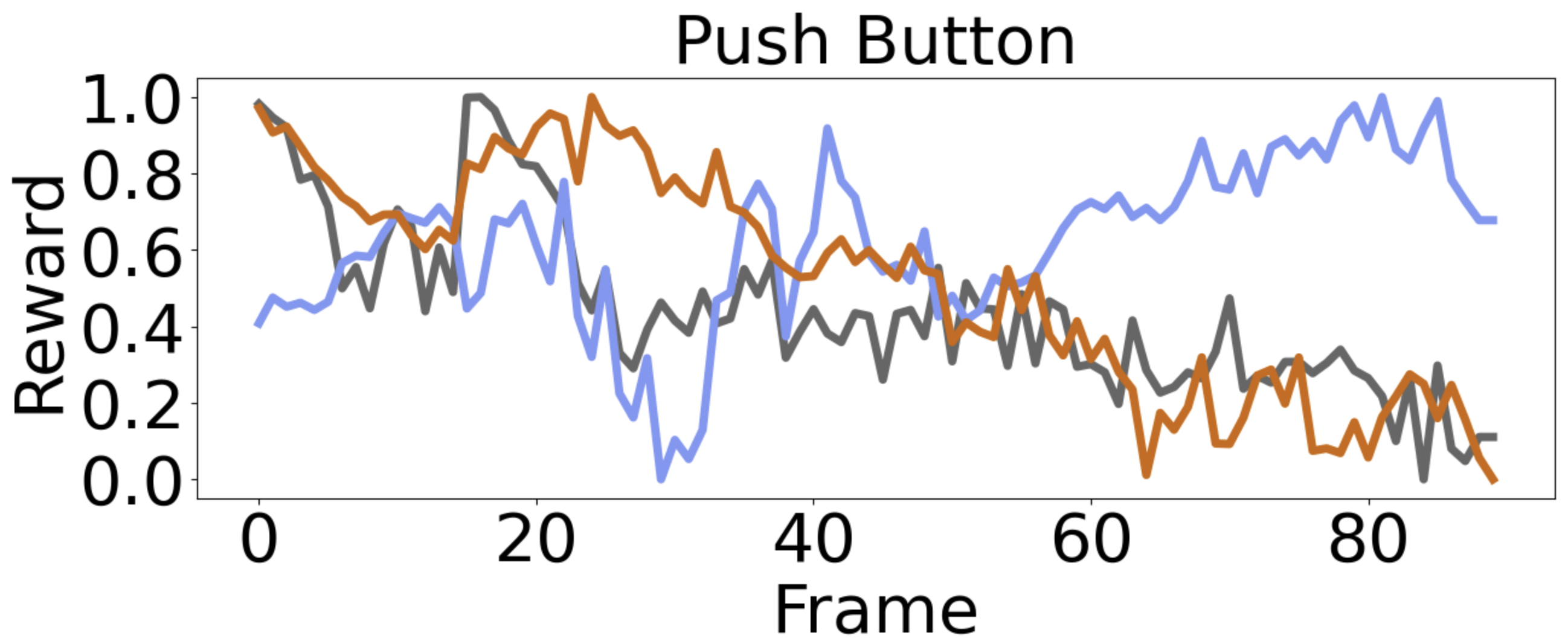}\hfill
    \includegraphics[width=0.33\linewidth]{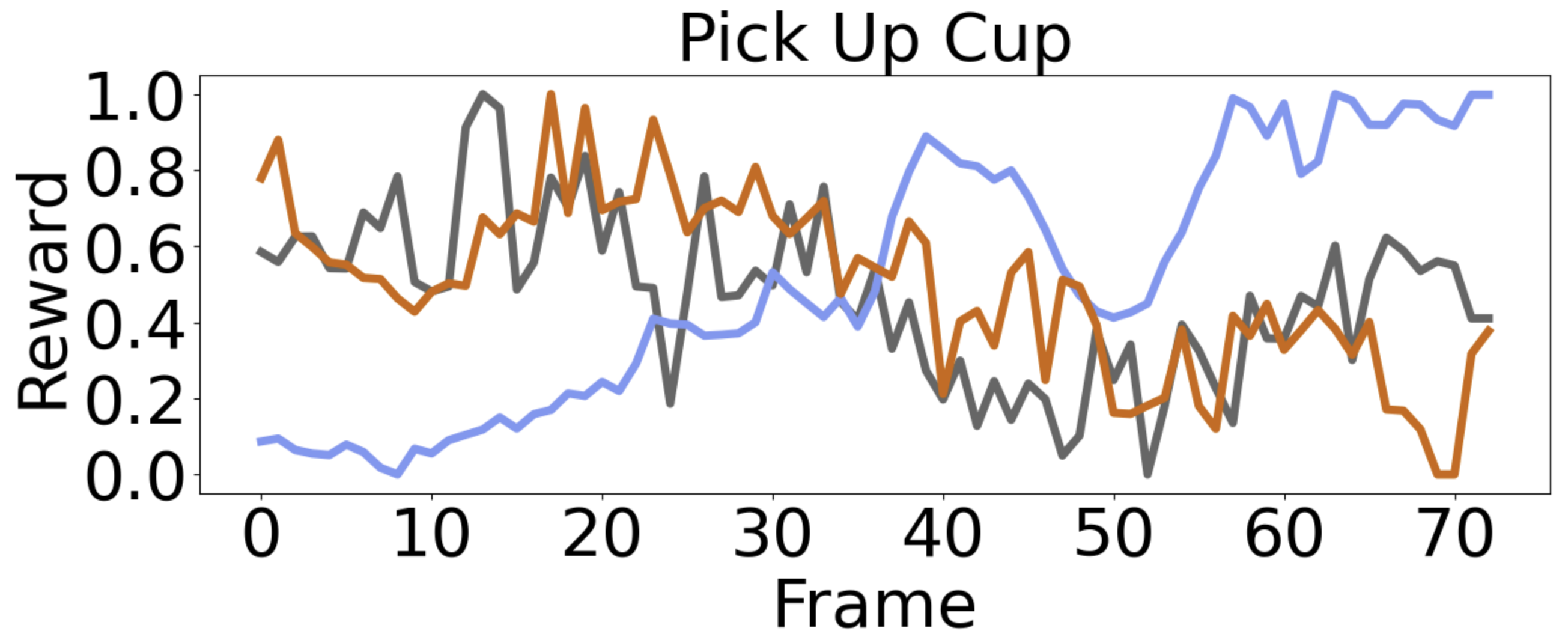}\par
    \vspace{10pt}
    \includegraphics[width=0.4\linewidth]{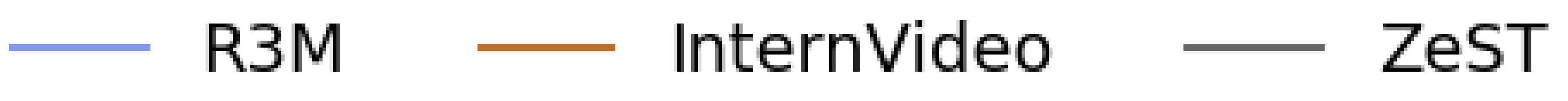}
    \caption{Video-Language alignment scores from R3M~\citep{r3m}, InternVideo~\citep{internvideo}, and ZeST~\citep{zest} on RLBench downstream tasks plotted over an expert episode with 3 snapshots visualized. Rewards are highly noisy and do not increase smoothly throughout the episode. Optimizing this signal with RL is unlikely to lead to stable solutions, and thus we instead use rewards as an exploration signal during pretraining.}
    \label{fig:expert}
\end{figure*}

\paragraph{R3M score as a reward}
To extract pretraining reward signals for RL, we propose to use R3M score as a reward.
Our motivation is that the R3M score is better suited for providing shaped rewards because its representations are explicitly trained to understand temporal information within videos (see Section~\ref{sec:background} for details) in contrast to other VLMs without such components \citep{clip,videoclip,internvideo}.
Specifically, we define our intrinsic reward using the R3M score as below:
\begin{equation} \label{eq:r3mscore}
     r^{\texttt{LAMP}}_{i} = G_{\theta}(F_{\phi}(o_{1}), F_{\phi}(o_{i}), L_{\alpha}(\mathbf{x}))
 \end{equation}
where $G_{\theta}$ denotes the score predictor in R3M.
Intuitively, this reward measures how $o_{i}$ is making a progress from $o_{1}$ towards achieving the tasks specified by natural language instruction $\mathbf{x}$.
We find that our intrinsic reward is indeed better aligned with the progress within expert demonstrations in our considered setups compared to other VLMs (see Figure~\ref{fig:expert} for supporting experiments).

\paragraph{Rewards with diverse language prompts}
To fully exploit the language understanding of VLMs, we propose to query them with a diversity of texts describing   a diversity of objectives, as opposed to computing the reward by repeatedly using a single instruction. Specifically, we obtain diverse, semantic rewards modulated by language, generating diverse sets of language instructions for each task and use them for prompting the model.
We query ChatGPT\footnote{\url{https://chat.openai.com}} to generate diverse language instructions of two types: imperative instructions and statements of completion (\textit{e.g.} move the mug vs. the mug is moved). Given that large-scale video datasets are predominantly human-centric, we obtain prompts that are human-centric, robot centric, as well as ambiguous (\textit{e.g.} the robot arm moved the mug vs. use your hand to move the mug vs. reach toward the mug and move it).
Moreover, we augment the instructions by querying for synonym nouns.
By inputting diverse language instructions from the dataset along with the agent's image observations, we effectively modulate the frozen, pretrained R3M reward and produce diverse semantic rewards that are grounded in the visual environment of the agent.

\subsection{Language-Conditioned Behavior Learning}
\label{sec:method_behavior}
While several recent works have shown that rewards from VLMs can be used for training RL agents~\citep{minedojo,zest}, it is still questionable whether these rewards can serve as a sufficiently reliable signal for inducing the intended behavior.
In this work, we instead propose to leverage such rewards from VLMs as a pretraining signal for RL policies, utilizing the knowledge captured within large VLMs for scripting diverse and useful exploration rewards.

\paragraph{Pretraining environment}
To learn diverse behaviors that can be transferred to various downstream tasks, we design a set of tasks with realistic visuals and diverse objects.
Specifically, we build a custom environment based on the RLBench simulation toolkit~\citep{rlbench}. In order to simulate a realistic visual scene, we download images from the Ego4D dataset~\citep{ego4d} and overlay them as textures on the tabletop and background of the environment (see Figure~\ref{fig:method}). To produce diverse objects and affordances, we import ShapeNet~\citep{shapenet} object meshes into the environment. Both the visual textures and the objects are randomized every episode of training.

\paragraph{Objective}
Because the LAMP reward can be seen as measuring the extent to which the agent is closer to solving the task (see Section~\ref{sec:method_reward}), it can be readily be combined with novelty-seeking unsupervised RL methods that optimize both extrinsic and intrinsic rewards. Therefore, to incentivize exploration, we combine the LAMP reward with the novelty score from a separate exploration technique. 
Specifically, we consider Plan2Explore~\citep{plan2explore} that utilizes the disagreement between future latent state predictions as a novelty score.
Let this novelty-based score be $r^{\texttt{P2E}}_{i}$. We then train our pretraining agent to maximize the following weighted sum of rewards:
\begin{equation} \label{eq:lamp}
    r^{\texttt{pre}}_{i} = \alpha \cdot r^{\texttt{P2E}}_{i} + (1-\alpha) \cdot r^{\texttt{LAMP}}_{i}
\end{equation}
where $\alpha$ is a hyperparameter that balances the two rewards. By combining this novelty-based reward with the LAMP reward, we encourage the agent to efficiently explore its environment but with an additional bias towards interacting with the semantically meaningful affordances. We found that an $\alpha$ value of 0.9 works quite well across the tasks evaluated.

\begin{wrapfigure}[18]{R}{0.35\textwidth}
\vspace{-1.5em}
\centering
    \begin{subfigure}[]{0.48\linewidth}
        \includegraphics[width=\linewidth]{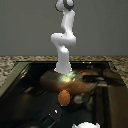}
     \end{subfigure}
     \begin{subfigure}[]{0.48\linewidth}
        \includegraphics[width=\linewidth]{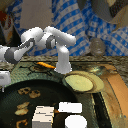}
     \end{subfigure}
     \begin{subfigure}[]{0.48\linewidth}
        \includegraphics[width=\linewidth]{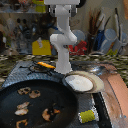}
     \end{subfigure}
     \begin{subfigure}[]{0.48\linewidth}
        \includegraphics[width=\linewidth]{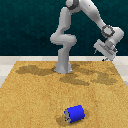}
     \end{subfigure}
     \caption{We pretrain on domain-randomized environments based on Ego4D textures, occasionally sampling the default, non-randomized RLBench environment.}
     \label{fig:env}
\end{wrapfigure}

\paragraph{Pretraining pipeline}
During task-agnostic pretraining, the agent is deployed in a language-conditioned MDP where there are no environment task rewards $r^{\texttt{e}}_{i}$. For the underlying RL algorithm, we use Masked World Models (MWM)~\citep{mwm}, an off-policy, model-based method with architectural inductive biases suitable for fine-grained robot manipulation. Every episode, our method randomly samples some language prompt $\mathbf{x}$ from the generated dataset as specified in Section~\ref{sec:method_reward}.
Then we condition the MDP and agent on the respective embedding $L_{\alpha}(\mathbf{x})$ such that each rolled out transition in the episode can be expressed as $(o_i, a_i, o_{i+1}, L_{\alpha}(\mathbf{x}))$.
After each episode is collected, we compute the rewards for each transition by embedding the observations with the R3M visual encoder $F_{\phi}$ and then applying the R3M score predictor $G_{\theta}$ - afterwards adding the data to the replay buffer.
We then sample batches of transitions from the buffer and make reinforcement learning updates to a language-conditioned policy, critic, and world model.
By the conclusion of pretraining, we obtain a language-conditioned policy capable of bootstrapping diverse behaviors specified by the language $\mathbf{x}$.

\subsection{Downstream Task Adaptation}
\label{sec:method_adaptation}

In order to evaluate the quality of the pretrained skills, we evaluate on downstream reinforcement learning tasks with scripted task rewards $r_i^{\text{e}}$. Since we have learned a language-conditioned policy, we simply select a language instruction $\mathbf{x_{ft}}$ roughly corresponding to the downstream task semantics in order to condition the pretrained agent. We remark that an additional advantage of LAMP is its use of language as a task-specifier, which enables this simplicity of zero-shot selection of a policy to finetune \citep{irm}. We fix this language instruction selection for the entirety of task learning and finetune all RL agent model components except the critic which we linear probe for training stability.

\begin{algorithm}[t!]
    \caption{Language Reward Modulated Pretraining (LAMP)}
    \begin{algorithmic}[1]
        \STATE Initialize Masked World Models (MWM) parameters 
        \STATE Load pretrained DistilBERT $L_{\alpha}$
        \STATE Load pretrained R3M visual encoder $F_{\phi}$
        \STATE Load pretrained R3M score predictor $G_{\theta}$
        \STATE Initialize replay buffer $\mathcal{B} \gets 0$
        \STATE Prefill language prompt buffer $\mathcal{B}^l$ 
        \STATE Prefill synonym buffer $\mathcal{B}^s$
        \FOR{each episode} 
            \STATE Randomize scene textures by sampling among Ego4D and original RLBench textures
            \STATE Sample ShapeNet Objects to place in scene
            \STATE Sample language prompt $\mathbf{x}$ from $\mathcal{B}^l$ (e.g., \texttt{Pick up the [NOUN]})
            \STATE Replace $\texttt{[NOUN]}$ in $\mathbf{x}$ by sampling a synonym from $\mathcal{B}^s$ for a random object in the scene
            \STATE Process the prompt via DistilBERT to obtain language embeddings $L_\alpha(\mathbf{x})$
            \STATE Collect episode transitions with $\pi(a|(s,L_{\alpha}(\mathbf{x})))$
            \STATE Assign LAMP rewards to all time steps (in parallel) by embedding frames with $F_{\phi}$ and querying the R3M score predictor $G_{\theta}$
            \STATE Add all episode transitions to $\mathcal{B}$
            \STATE Update MWM and Plan2Explore parameters as in \citep{mwm, plan2explore} by sampling transitions from $\mathcal{B}$ and augmenting LAMP rewards with novelty bonus to train agent
        \ENDFOR
        
    \end{algorithmic}
    \label{alg:lamp}
\end{algorithm}

\section{Experiments}

\subsection{Setup}

\paragraph{Environment details}
As previously mentioned in Section~\ref{sec:method_behavior}, we consider domain-randomized environments for pre-training (see Figure~\ref{fig:env} for examples).

Specifically, our pre-training environments consist of 96 domain-randomized environments with different Ego4D textures overlayed over the table, walls, and floor.
We also sample the environments having default RLBench environment textures with probability of 0.2.

For finetuning, we implement a shaped reward function based on the ground truth simulator state and train the agent to optimize this signal instead of the pretraining reward.
We use the exact scenes released in RLBench in order to encourage reproducibility, notably keeping the default background and table textures fixed throughout the course of training.
We use a 4-dimensional continuous action space where the first three dimensions denote end-effector positional displacements and the last controls the gripper action. We fix the end-effector rotation and thus select tasks that can be solved without changing the orientation of the end-effector.

\paragraph{Baselines}
As a baseline, we first consider a randomly initialized MWM agent trained from scratch to evaluate the benefit of pretraining. In order to evaluate the benefit of pretraining with LAMP that modulates the reward with language, we also consider Plan2Explore~\citep{plan2explore} as a baseline, which is a novelty-seeking method that explores based on model-uncertainty.

\begin{figure*}[t!]
\centering
    \begin{subfigure}[]{0.32\textwidth}
        \includegraphics[width=\textwidth]{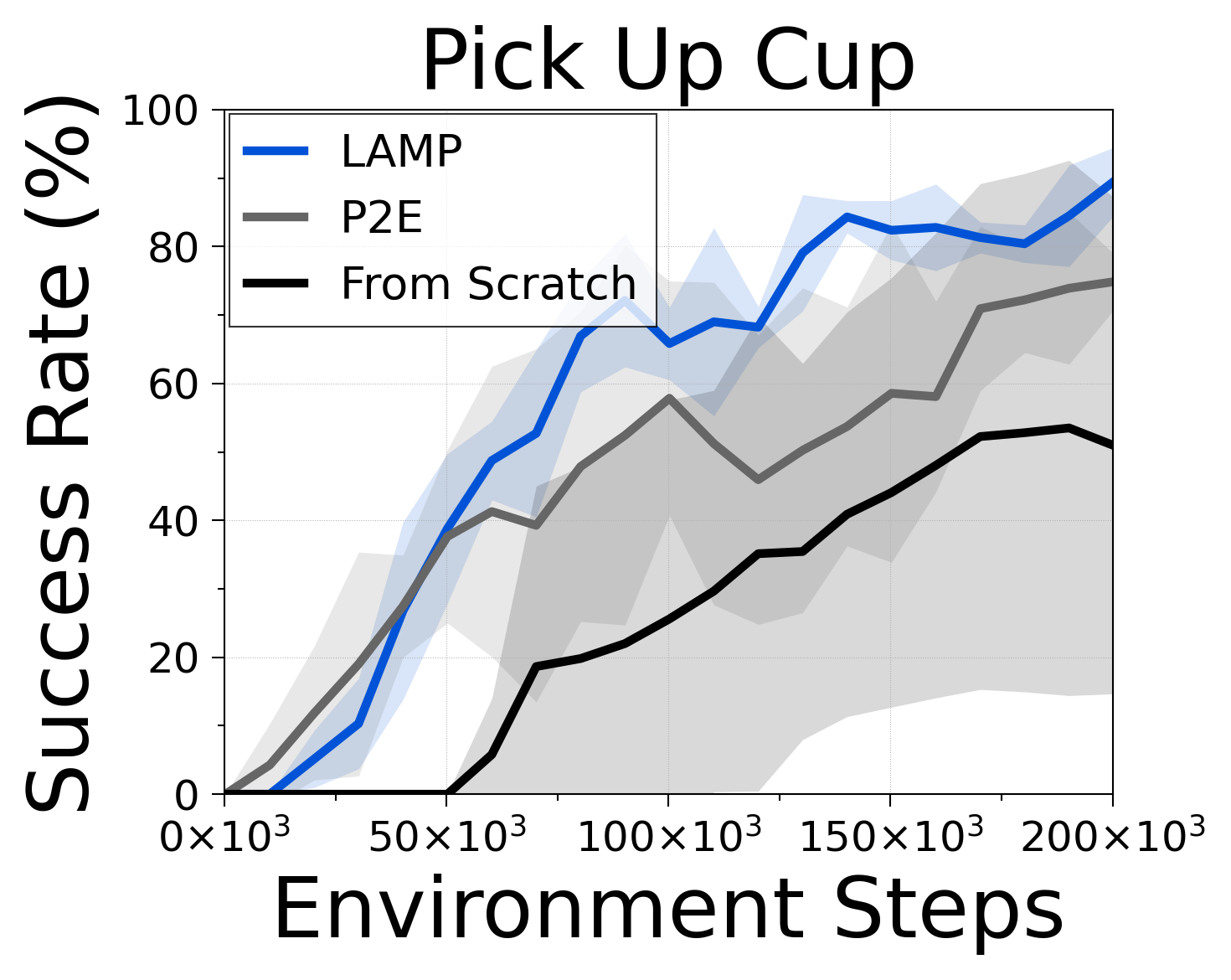}
     \end{subfigure}
     \begin{subfigure}[]{0.32\textwidth}
        \includegraphics[width=\textwidth]{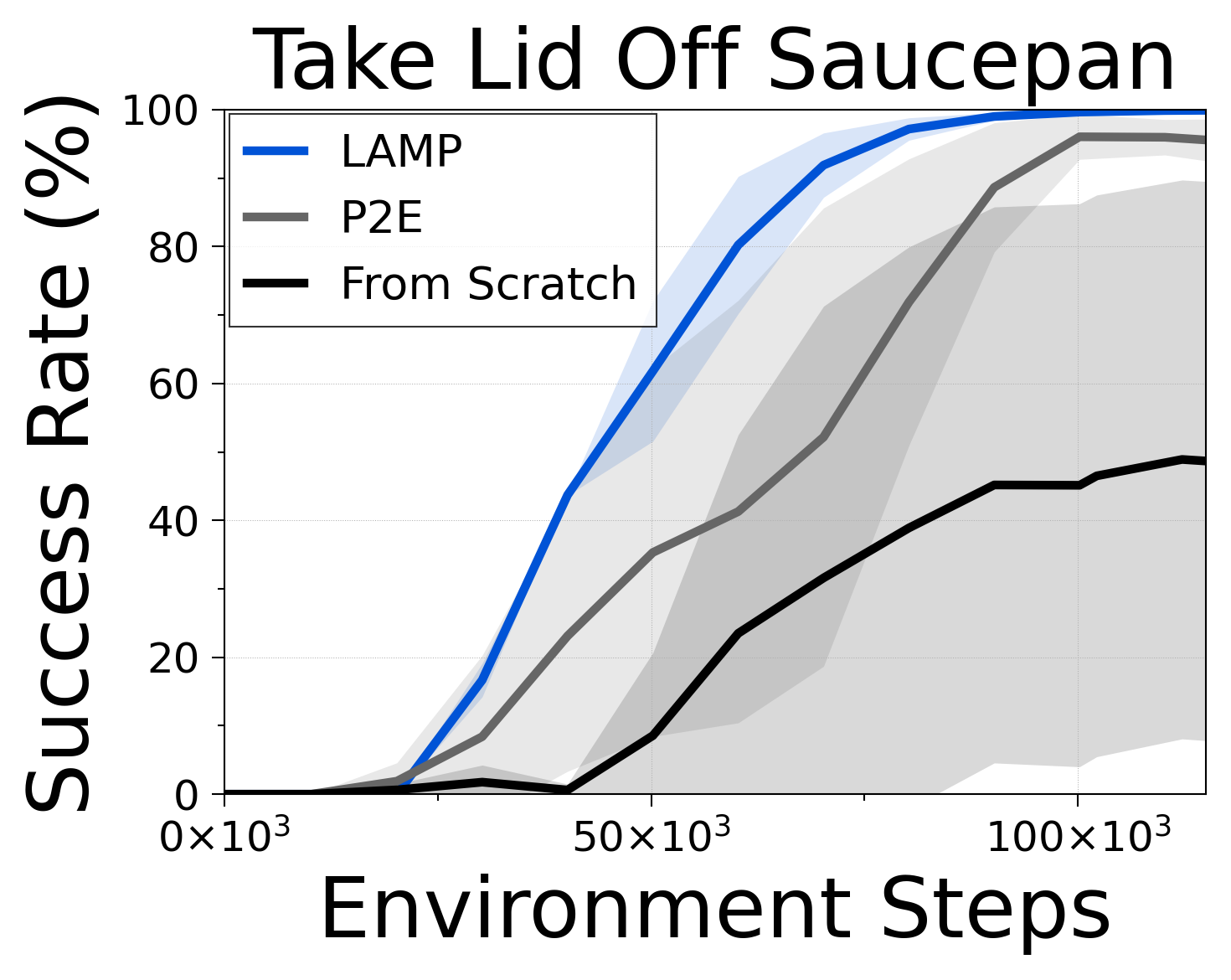}
     \end{subfigure}
     \begin{subfigure}[]{0.32\textwidth}
        \includegraphics[width=\textwidth]{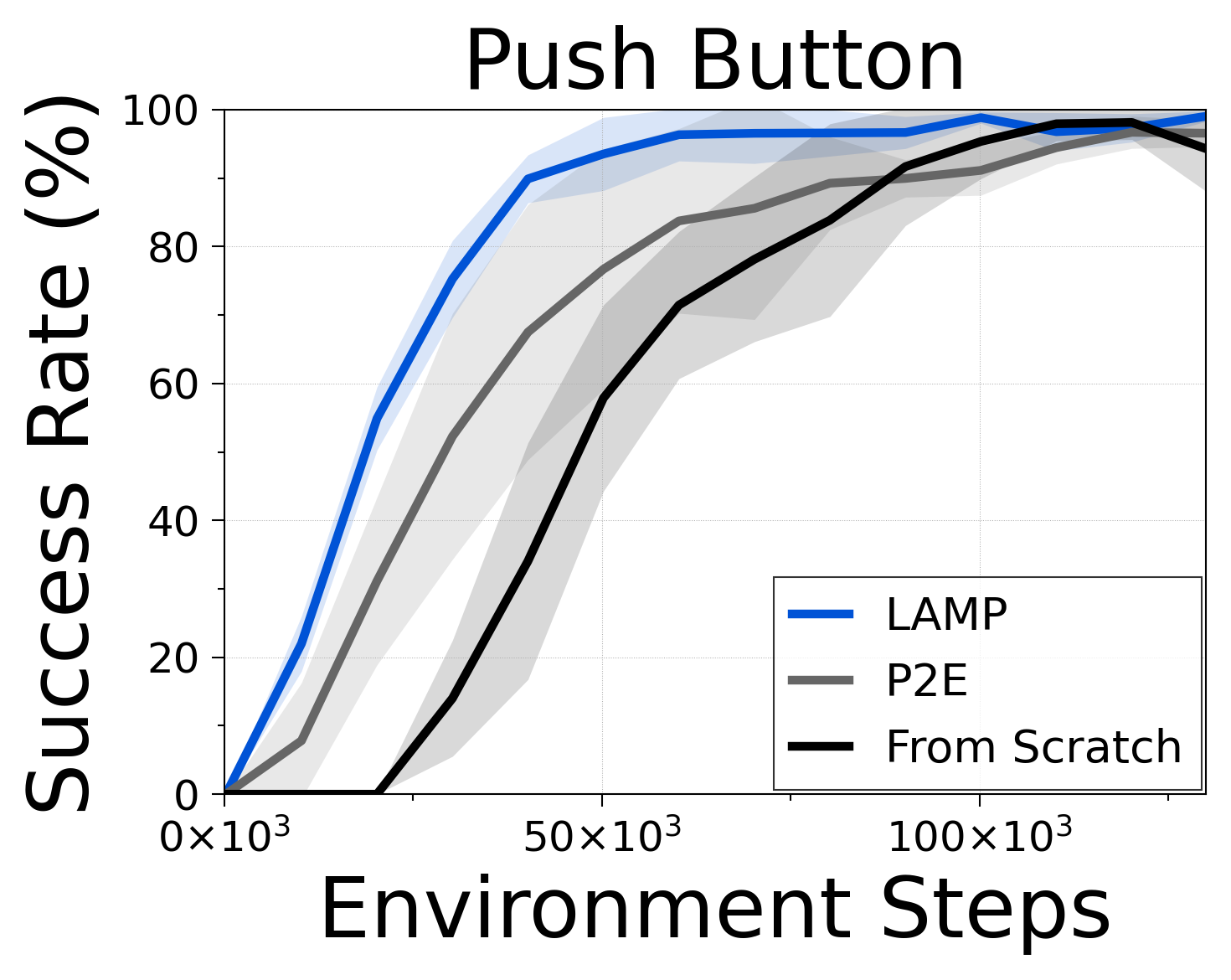}
     \end{subfigure}
     \begin{subfigure}[]{0.32\textwidth}
        \includegraphics[width=\textwidth]{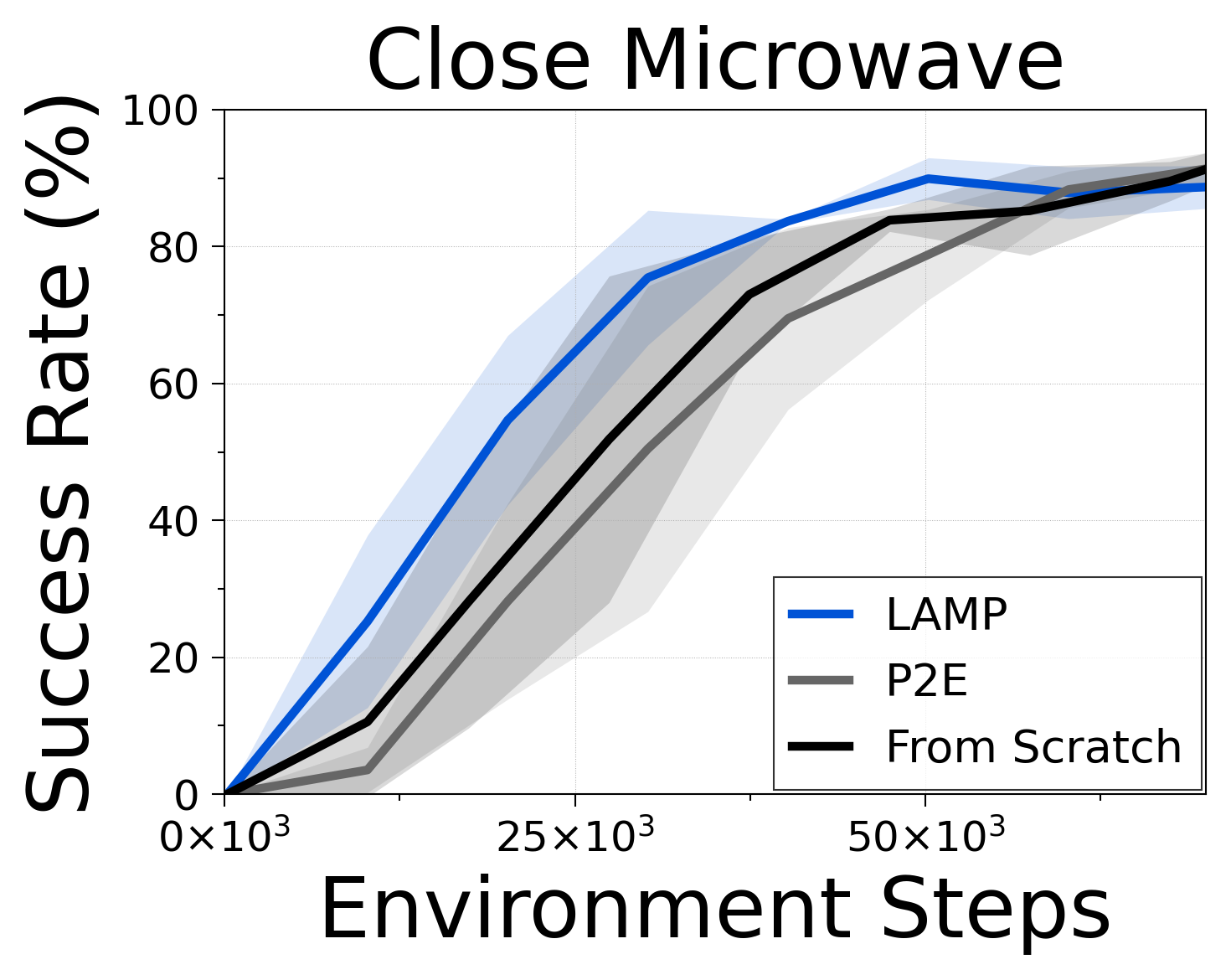}
     \end{subfigure}
     \begin{subfigure}[]{0.32\textwidth}
        \includegraphics[width=\textwidth]{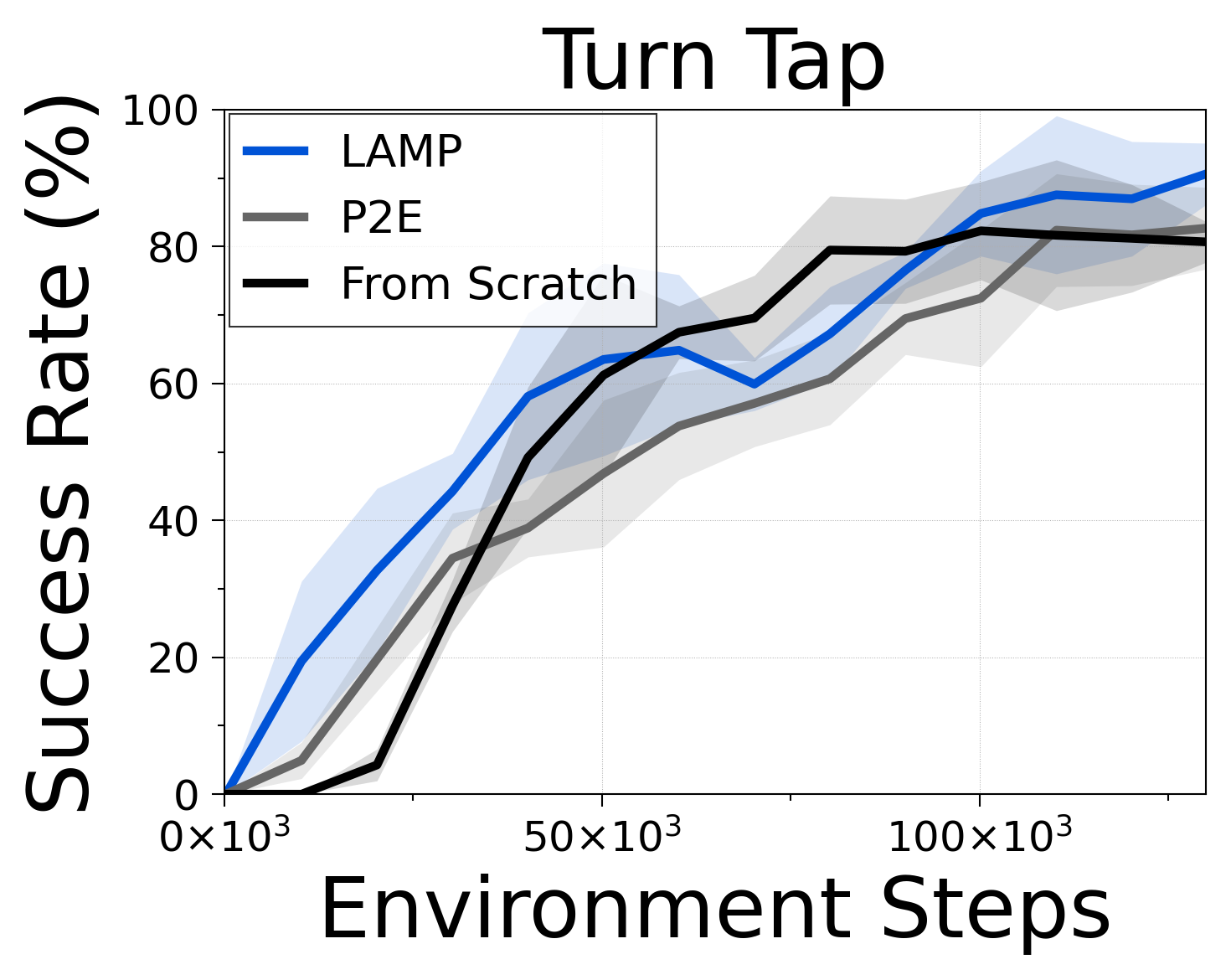}
     \end{subfigure}
     \caption{Finetuning performance on visual robotic manipulation tasks in RLBench. We provide the performance on additional tasks in the supplementary material. The solid line and shaded region represent mean and standard deviation across 3 seeds.}
     \label{fig:finetune}
\end{figure*}

\subsection{Results}
Across tasks in Figure~\ref{fig:finetune}, training to a new task from a randomly initialized agent from scratch exhibits high-sample complexity in order to learn a performant policy. Across most RLBench tasks, Plan2Explore, which employs purely unsupervised exploration, exceeds the performance of training from scratch. LAMP outperforms or is competitive with Plan2Explore and consistently outperforms training from scratch. We hypothesize that this is because by pretraining with a VLM reward, LAMP biases the agent to explore efficiently in ways that are semantic meaningful and avoids spending time investigating spurious novelty. It is also possible that by optimizing more diverse rewards during pretraining, the agent learns representations that allow it to quickly adapt to the unseen task reward during finetuning.

\section{Ablations}
\label{ablations}

\subsection{Language Prompting}
\label{langprompt}
A particular exciting benefit of using VLMs trained on internet-scale data is the diversity of language we can query for plausibly infinite rewards. We ablate different prompt styles used during the pretraining phase. The 6 language prompting styles are as such:
\begin{enumerate}
\item \textbf{Prompt Style 1}: \texttt{Pick up the [NOUN].} 
\item \textbf{Prompt Style 2}: \texttt{[RELEVANT VERB] and [SYNONYM NOUN].}
\item \textbf{Prompt Style 3}: \texttt{[RELEVANT VERB] and [RANDOM NOUN].}
\item \textbf{Prompt Style 4}: \texttt{[IRRELEVANT VERB] and [SYNONYM NOUN].}
\item \textbf{Prompt Style 5}: \texttt{[IRRELEVANT VERB] and [RANDOM NOUN].}
\item \textbf{Prompt Style 6}: \texttt{[Snippets from Shakespeare].}
\end{enumerate}

For Prompt Styles 1-5, we compare the effect of using relevant and irrelevant nouns and verbs, though all remain action-based and task-relevant. For Prompt Style 6, we select snippets of Shakespeare text to see the effect of pretraining on rewards generated from completely out of distribution and implausible tasks.

The prompting styles provide varying types of language diversity during pretraining. We evaluate how important it is that the prompts be aligned with possible tasks in the environment-- might an instruction like "sweep the floor" still encourage possibly more interesting behaviors even when the agent is only presented with mugs? In addition, providing a multitude of prompts may mitigate the adverse effects of overfitting by exploiting the imperfections of the VLM rewards for a particular prompt.

\begin{figure*}[t!]
\centering
    \begin{subfigure}[]{0.32\textwidth}
        \includegraphics[width=\textwidth]{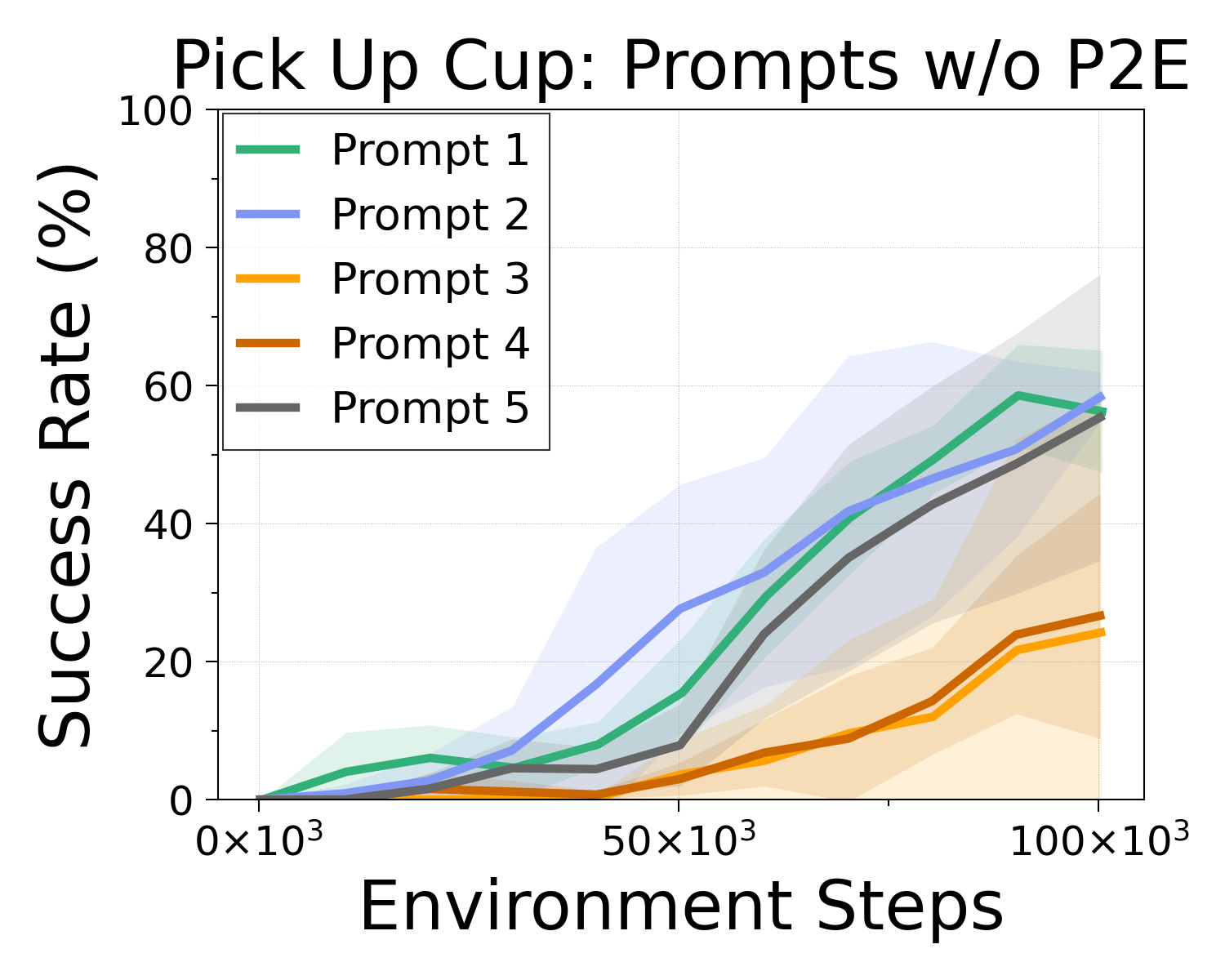}
     \end{subfigure}
     \begin{subfigure}[]{0.32\textwidth}
        \includegraphics[width=\textwidth]{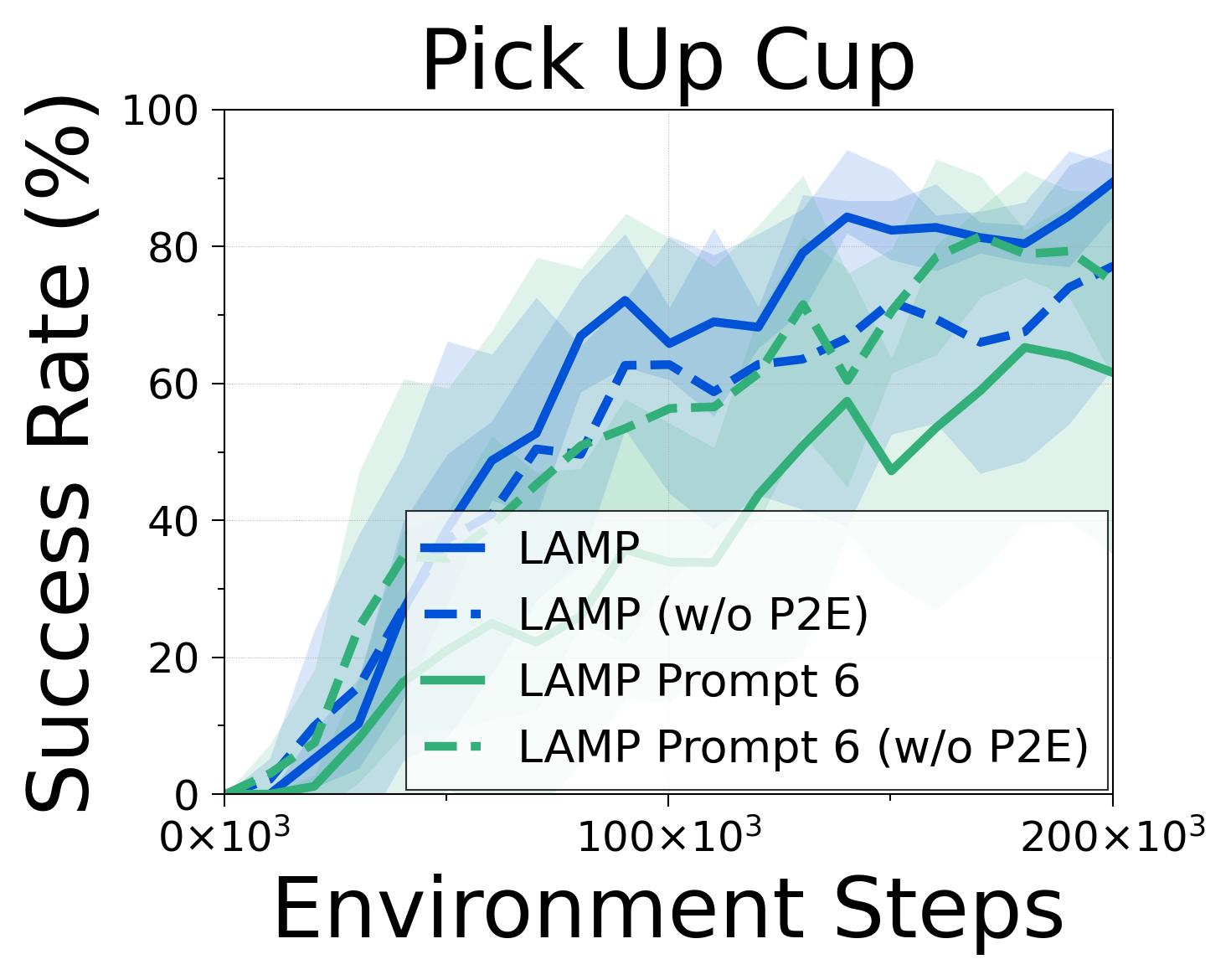}
     \end{subfigure}
     \caption{Finetuning performance on RLBench tasks. (a) Effect of pretraining with rewards from different language prompting styles. Language prompts focus on action-based tasks. (b) Effect of pretraining on action-based prompts (Lang 2) and random prompts (Lang 6).}
     \label{fig:lang}
\end{figure*}

In Figure~\ref{fig:lang}, we compare the finetuning results of Prompts 1-5, which are action-based prompts. We choose the task "Pick Up Cup" because the task name is simple and similar to prompts during pretraining. We find that for this task, Prompt Style 2, which pretrains on semantically similar but diverse wordings of prompts, is most successful. In addition, Prompt Style 1, which pretrains on very simple instructions, finetunes efficiently, as well. For our main experiments, we choose Prompt 2 based on its strong finetuning performance.

In Figure~\ref{fig:lang}, we also compare the performance of our best action-based prompt, Prompt 2, with a non-action-based prompt, Prompt 6, combined with auxiliary exploration objectives. While LAMP Prompt 6 (w/o Plan2Explore) and LAMP Prompt 2 (w/o Plan2Explore) perform similarly, we notice that adding in the exploration objectives dramatically decreases the finetuning effectiveness of LAMP Prompt 6. We hypothesize that both exploration coverage and exploration efficiency during pretraining play an important role. By separately increasing exploration coverage through Plan2Explore, the quality of the VLM rewards may become more important for focusing the auxiliary exploration objective. Thus, LAMP Prompt 2, which incorporates Plan2Explore, is trained on higher-quality, more relevant rewards, and can explore more efficiently during pretraining, and therefore can finetune more effectively.

Overall, the differences in pretraining with different prompting styles is not extreme, suggesting that LAMP is robust to different prompting strategies, and providing diverse instructions can be an effective way of pretraining an agent.

\subsection{VLM Model}

\begin{wrapfigure}[11]{R}{0.32\textwidth}
\vspace{-5.5em}
\centering
    \begin{subfigure}[]{\linewidth}
        \includegraphics[width=\linewidth]{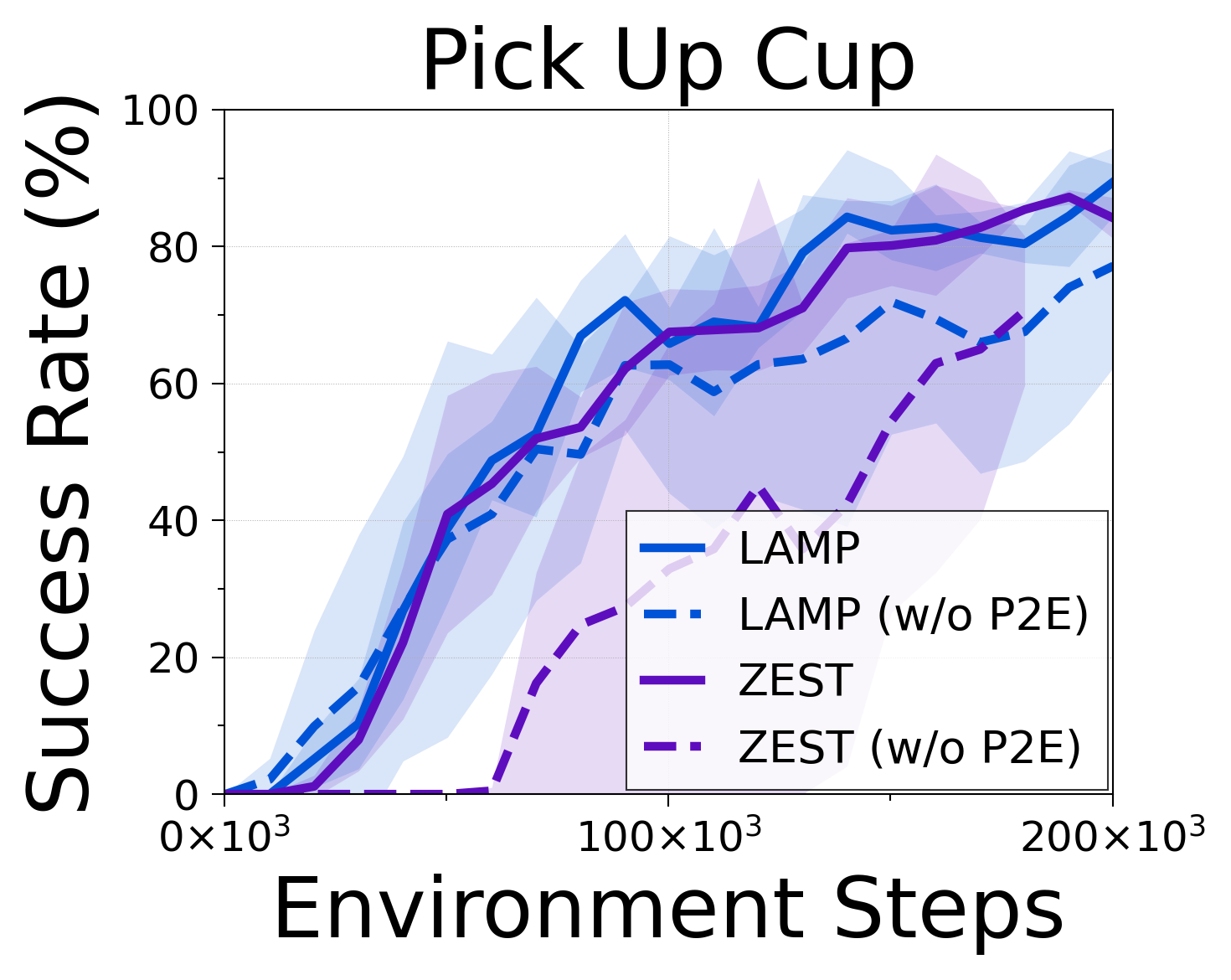}
     \end{subfigure}
     \caption{Finetuning performance for models pretrained with rewards extracted from different VLMs.}
     \label{fig:vlm}
\end{wrapfigure}

We compare our method across different vision-language reward models. CLIP, which is trained on static image-caption pairs, can serve as a reward model by extracting the cosine similarity between text feature and image feature changes as presented as ZeST in \citep{zest}. Following ZeST, in order to express feature displacements, we assign context features in image and language space as $s_0$, the initial image in an episode, and $\mathbf{x_0}$, a language prompt that inverts the action (open microwave inverts close microwave) described in the desired instruction $\mathbf{x}$. In particular, we use the following reward parameterization, 
\begin{equation}
    r_i = (F_{\phi}(s_i)-F_{\phi}(s_0)) \cdot (L_{\alpha}(\mathbf{x}) - L_{\alpha}(\mathbf{x_0})).
\end{equation}
where $F_{\phi}$ is the CLIP pretrained image encoder, $L_{\alpha}$ is the CLIP pretrained language encoder, and the task reward $r_i$ is defined as the dot product of the visual and language delta embeddings.

We show finetuning results on the RLBench Pick Up Cup downstream task in Figure \ref{fig:vlm}. While R3M seems to lead to consistently good performance, ZeST pretraining can also perform reasonably well. From this we conclude that our method is not inherently reliant on a particular VLM and is likely to benefit from more powerful VLMs in the future.

\section{Discussion}
In this work, we present LAMP, an algorithm for leveraging frozen VLMs to pretrain reinforcement learning agents. We observe a number of limitations of using LRFs for task reward specifcation in RL, and propose a new setting and methodology that is more accommodating of these limitations. We demonstrate that by leveraging the flexibility and zero-shot generalization capabilities of VLMs, we can easily produce diverse rewards during pretraining that encourage the learning of semantically grounded exploratory behaviors. Furthermore, we show that VLM parameterized rewards exhibit strong synergies with novelty-seeking RL pretraning methods and can lead to strong performance in combination with Plan2Explore. We evaluate LAMP on a number of challenging downstream tabletop manipulation tasks from RLBench as well as show evidence that our method is likely to benefit from more capable VLMs. Overall, we see LAMP as an indication of the promise of leveraging large pretrained VLMs for pretraining behaviors in challenging environments. A limitation of LAMP is its reliance on performing inference of the VLM model many times throughout the course of pretraining in order to generate rewards. Inference speed for larger, more powerful VLMs may be slow, bottlenecking the speed of the pretraning process. Another limitation is that LAMP does not address the long-horizon sequencing setting where we might be interested in finetuning agents conditioned on many language prompts. We leave this as in intriguing direction for future work. \\

\newpage
\bibliography{neurips_2023}
\bibliographystyle{plain}

\newpage
\appendix
\onecolumn

\section{Masked World Models}
For our experiments, we use Masked World Models (MWM) \citep{mwm} as our underlying RL algorithm.
MWM is a model-based RL framework that aims to decouple visual representation learning and dynamics learning.
In contrast to prior algorithms that learn world models in an end-to-end manner, MWM separately learns a visual encoder via masked autoencoding~\citep{he2021masked} and a world model that reconstructs frozen autoencoder representations.
We build LAMP upon the official implementation provided by authors (\url{https://github.com/younggyoseo/MV-MWM}) which supports experimentation on RLBench~\citep{rlbench}.
In Table \ref{tab:maetable}, \ref{tab:wmtable}, and \ref{tab:actorcritictable} we provide all relevant hyperparameters.

\section{Video-Language Models}
\paragraph{R3M} For the R3M \citep{r3m} VLM in LAMP we use the official implementation (\url{https://github.com/facebookresearch/r3m}). We use the ResNet-18 visual backbone and preserve all default hyperparameters. To encode the language prompts, we use the DistilBERT `base-uncased' model (\url{https://huggingface.co/distilbert-base-uncased}) from the transformers (\url{https://pypi.org/project/transformers/}) package as used in the R3M implementation.
\paragraph{InternVideo} For the InternVideo \citep{internvideo} VLM in LAMP we use the official implementation (\url{https://github.com/OpenGVLab/InternVideo}). We use the "B/16" model and pretrained weights provided by the authors for embedding the images and text.

For InternVideo \citep{internvideo} we match the style of alignment score computation used in training and inference. We use the following reward parameterization:
\begin{equation}
    r_i = F_{\phi}([s_{i//8}, s_{2*i//8}, ..., s_{8*i//8}])\cdot L_{\alpha}(l).
\end{equation}
where 8 frames evenly spaced from the agent's entire history are featurized by the visual encoder $F_{\phi}$.

\paragraph{ZeST} For the CLIP \citep{clip} VLM used in ZeST \citep{zest} for LAMP, we use the official OpenAI CLIP model (\url{https://github.com/openai/CLIP}). We use the "ViT-B/32" release of the CLIP visual encoder for embedding images and the CLIP text encoder for embedding langauge prompts.

\section{Experimental Details}
\subsection{Language Prompting Types}
We ablate on 6 different prompt styles, with the structures defined in Section~\ref{langprompt}. To construct Prompt Style 1, we replace the \texttt{[NOUN]} in the \texttt{Pick up the [NOUN].} prompt with the ShapeNet name. To construct Prompt Style 6, we sample from random Shakespeare phrases listed below. For the remaining Prompt Styles, we sample a verb structure, either \texttt{IRRELEVANT} or \texttt{RELEVANT}; and a noun, either \texttt{RANDOM} or \texttt{SYNONYM}. Both verb structures are included in Table~\ref{tab:verb}. We include examples of random nouns and synonym nouns for a sample object; nouns and synonyms for all objects are ommitted for space, but they can be found in the code.

Verb structures and synonyms were curated from ChatGPT, with filtering afterward to select most relevant verbs. Random nouns were taken from the synonyms. We provide the automatically generated language prompt datasets for each language prompting scheme used during LAMP pretraining. Prompts to ChatGPT are included in Table~\ref{tab:chatgpt}.

\begin{table}[htbp]
  \centering
  \caption{\textbf{Using an LLM to generate verb structures and nouns}. We query ChatGPT with prompts to create a set of diverse tasks.}
  \label{tab:chatgpt}
  \begin{tabular}{|p{0.5\linewidth}|p{0.5\linewidth}|}
    \hline
    \textbf{Generating Verb Structures} & \textbf{Generating Synonyms} \\
    \hline
    Give me a list of 40 task variations that present an interesting task for a person to do in a home or kitchen scenario. 

    Examples should not be complicated, and should be possible to do very quickly, within a minute or so. These should be simple tasks that are interesting and diverse, but EASY. Tasks should be atomic and very general.

    For example:
    
    1. Reach for the mug
    
    2. Open the microwave
    
    3. Wipe the table clean
    
    4. Water the flowers 
    &
    Please give me 40 synonyms for bowl
    
    Example:
    
    1. bowl 
    
    2. soup bowl
    
    3. dish \\
    \hline
    
  \end{tabular}
\end{table}

\begin{table}[htbp]
  \centering
  \caption{\textbf{Verb Structures}. We include the different verb structures used during pretraining.}
  \label{tab:verb}
  \begin{tabular}{|l|l|}
    \hline
    \textbf{Relevant Verb} & \textbf{Irrelevant Verb} \\
    \hline
    \rowcolor[gray]{0.95}
    Pick up the [NOUN] &
    The [NOUN] is seized \\
    Lift the [NOUN] with your hands &
    The [NOUN] is clutched \\
    \rowcolor[gray]{0.95} Hold the [NOUN] in your grasp & 
    The [NOUN] is gripped \\
    Take hold of the [NOUN] and raise it &
    The [NOUN] is firmly grasped \\
    \rowcolor[gray]{0.95} Grasp the [NOUN] firmly and lift it up & 
    The [NOUN] is tightly held \\
    Raise the [NOUN] by picking it up &
    The [NOUN] is firmly caught \\
    \rowcolor[gray]{0.95} Retrieve the [NOUN] and hold it up & 
    The [NOUN] is securely clasped \\
    Lift the [NOUN] by gripping it &
    The [NOUN] is rotated \\
    \rowcolor[gray]{0.95} Seize the [NOUN] and raise it off the surface & 
    The [NOUN] has been flipped \\
    Hold onto the [NOUN] and lift it up &
    The [NOUN] has been knotted \\
    \rowcolor[gray]{0.95} The [NOUN] is lifted up & 
    The [NOUN] has been folded \\
    The [NOUN] is picked up off the ground &
    The [NOUN] has been rinsed \\
    \rowcolor[gray]{0.95} The [NOUN] is raised up by hand & 
    The [NOUN] has been filled \\
    The [NOUN] is grasped and lifted up &
    The [NOUN] is shaken \\
    \rowcolor[gray]{0.95} The [NOUN] is taken up by hand &
    The [NOUN] has been scooped \\
    The [NOUN] is retrieved and lifted up &
    The [NOUN] is poured \\
    \rowcolor[gray]{0.95} The [NOUN] is lifted off its surface & 
    The [NOUN] has been scrubbed \\
    The [NOUN] is elevated by being picked up &
    The [NOUN] is tilted \\
    \rowcolor[gray]{0.95} The [NOUN] is hoisted up by hand & 
    The [NOUN] has been heated \\
    The [NOUN] is scooped up and lifted &
    Reach for the [NOUN] \\
    \rowcolor[gray]{0.95} The [NOUN] is lifted by the hand &
    Grasp at the [NOUN] \\
    The [NOUN] is grasped and picked up &
    Stretch out to touch the [NOUN] \\
    \rowcolor[gray]{0.95} The [NOUN] is raised by the palm & 
    Move your arm towards the [NOUN] \\
    The [NOUN] is taken up by the fingers &
    Use the gripper to rinse the [NOUN] \\
    \rowcolor[gray]{0.95} The [NOUN] is held and lifted up &
    Position the end effector to fold the [NOUN] \\
    The [NOUN] is lifted off the surface by the arm &
    Reach out the robotic arm to wipe the [NOUN] \\
    \rowcolor[gray]{0.95} The [NOUN] is picked up and held by the wrist &
    Utilize the gripper to seize the [NOUN] \\
    The [NOUN] is scooped up by the palm and lifted &
    Guide the robotic arm to obtain the [NOUN] \\
    \rowcolor[gray]{0.95} The [NOUN] is elevated by the hand &
    Maneuver the end effector to lift up the [NOUN] \\
    The [NOUN] is taken up by the fingers of the hand &
    Extend your hand towards the [NOUN] \\
    \rowcolor[gray]{0.95} The [NOUN] is grasped and raised & 
    Reach out your hand to acquire the [NOUN] \\
    The [NOUN] is lifted by the gripper &
    Guide your arm to rotate the [NOUN] \\
    \rowcolor[gray]{0.95} The end effector picks up the [NOUN] &
    Maneuver your hand to shake up the [NOUN] \\
    The arm lifts the [NOUN] &
    Flip the [NOUN] \\
    \rowcolor[gray]{0.95} The [NOUN] is held aloft by the robotic hand & 
    Tap the [NOUN] \\
    The robotic gripper secures the [NOUN] &
    Fold the [NOUN] \\
    \rowcolor[gray]{0.95} The [NOUN] is lifted off the surface by the robotic arm &
    Rotate the [NOUN] \\
    The robotic manipulator seizes and elevates the [NOUN] &
    Brush the [NOUN] \\
    \rowcolor[gray]{0.95} The robotic end effector clasps and hoists the [NOUN] &
    Twist the [NOUN] \\
    The [NOUN] is taken up by the robotic gripper &
    Wipe the [NOUN] \\
    \hline
    
  \end{tabular}
\end{table}

\begin{table}[htbp]
\centering
\caption{\textbf{Prompt Style 6}. Snippets from Shakespeare.}
\label{tab:title}
\begin{tabular}{|p{\textwidth}|}
\hline
\rowcolor[gray]{0.95} Holla, Barnardo. \\
BARNARDO Say, what, is Horatio there? \\
\rowcolor[gray]{0.95} HORATIO A piece of him. \\
Welcome, Horatio.—Welcome, good Marcellus. \\
\rowcolor[gray]{0.95} HORATIO \\
What, has this thing appeared again tonight? \\
\rowcolor[gray]{0.95} BARNARDO I have seen nothing. \\
MARCELLUS \\
\rowcolor[gray]{0.95} Horatio says ’tis but our fantasy \\
And will not let belief take hold of him \\
\rowcolor[gray]{0.95} Touching this dreaded sight twice seen of us. \\
Therefore I have entreated him along \\
\rowcolor[gray]{0.95} With us to watch the minutes of this night, \\
That, if again this apparition come, \\
\rowcolor[gray]{0.95} He may approve our eyes and speak to it. \\
Tush, tush, ’twill not appear. \\
\rowcolor[gray]{0.95} Sit down awhile, \\
How now, Horatio, you tremble and look pale. \\
\rowcolor[gray]{0.95} Is not this something more than fantasy? \\
What think you on ’t? \\
\rowcolor[gray]{0.95} At least the whisper goes so: our last king, \\
Whose image even but now appeared to us, \\
\rowcolor[gray]{0.95} Was, as you know, by Fortinbras of Norway, \\
Thereto pricked on by a most emulate pride, \\
\rowcolor[gray]{0.95} Dared to the combat; in which our valiant Hamlet \\
(For so this side of our known world esteemed him) \\
\rowcolor[gray]{0.95} Did slay this Fortinbras, who by a sealed compact, \\
Well ratified by law and heraldry, \\
\rowcolor[gray]{0.95} Did forfeit, with his life, all those his lands \\
Which he stood seized of, to the conqueror. \\
\rowcolor[gray]{0.95} Against the which a moiety competent \\
Was gagèd by our king, which had returned \\
\rowcolor[gray]{0.95} To the inheritance of Fortinbras \\
Had he been vanquisher, as, by the same comart \\
\rowcolor[gray]{0.95} And carriage of the article designed, \\
His fell to Hamlet. Now, sir, young Fortinbras, \\
\rowcolor[gray]{0.95} Of unimprovèd mettle hot and full, \\
Hath in the skirts of Norway here and there \\
\rowcolor[gray]{0.95} Sharked up a list of lawless resolutes \\
BARNARDO \\ 
\hline
\end{tabular}
\end{table}

\begin{table}[htbp]
  \centering
  \caption{\textbf{Nouns}. Synonym and random nouns for "bag."}
  \label{tab:verb}
  \begin{tabular}{|l|l|}
    \hline
    \textbf{Synonym Noun} & \textbf{Random Noun} \\
    \hline
    \rowcolor[gray]{0.95}
    bag &
    cap \\
    handbag &
    hat \\
    \rowcolor[gray]{0.95} purse & 
    snapback \\
    clutch &
    faucet \\
    \rowcolor[gray]{0.95} tote & 
    tap \\
    backpack &
    vase \\
    \rowcolor[gray]{0.95} knapsack & 
    flask \\
    satchel &
    earphone \\
    \rowcolor[gray]{0.95} shoulder bag & 
    earpiece \\
    duffel bag &
    knife \\
    \rowcolor[gray]{0.95} messenger bag & 
    blade \\
    grip &
    laptop \\
    \rowcolor[gray]{0.95} briefcase & 
    notebook \\
    pouch &
    vase \\
    \rowcolor[gray]{0.95} fanny pack &
    flowerpot \\
    drawstring &
    telephone \\
    \rowcolor[gray]{0.95} beach bag & 
    flip phone \\
    grocery shop &
    handle\\
    \rowcolor[gray]{0.95} shopping bag & 
    lever \\
    gift bag &
    gift bag \\
    \rowcolor[gray]{0.95} lunch bag &
    lunch  bag \\
    laptop bag &
    laptop bag \\
    \rowcolor[gray]{0.95} travel bag & 
    travel bag \\
    \hline
    
  \end{tabular}
\end{table}

\subsection{ShapeNet Objects}
\begin{figure*}[]
\centering
     \begin{subfigure}[]{0.15\textwidth}
        \includegraphics[width=\textwidth]{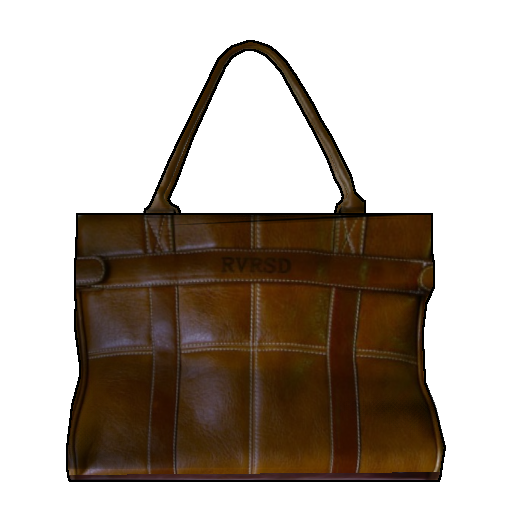}
     \end{subfigure}
     \begin{subfigure}[]{0.15\textwidth}
        \includegraphics[width=\textwidth]{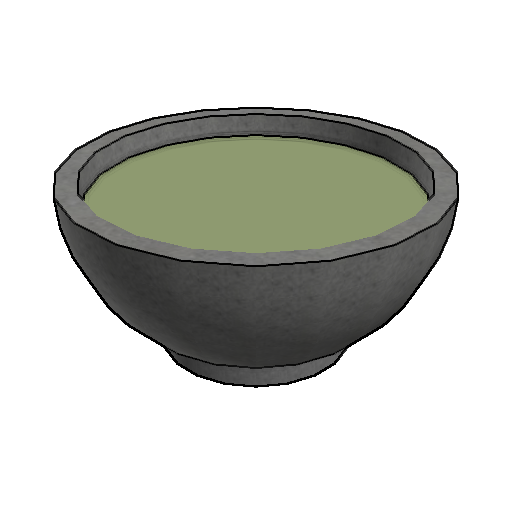}
     \end{subfigure}
     \begin{subfigure}[]{0.15\textwidth}
        \includegraphics[width=\textwidth]{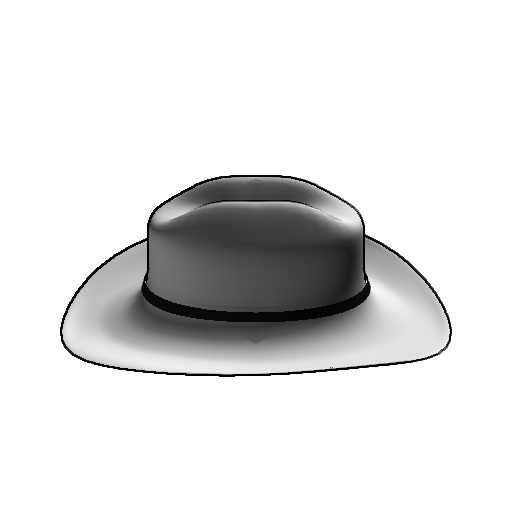}
     \end{subfigure}
     \begin{subfigure}[]{0.15\textwidth}
        \includegraphics[width=\textwidth]{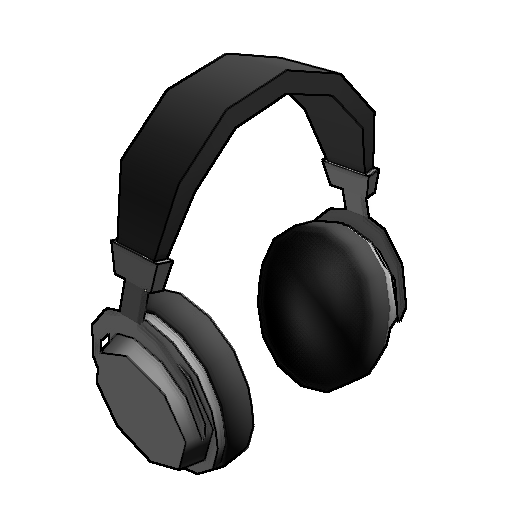}
     \end{subfigure}
     \begin{subfigure}[]{0.15\textwidth}
        \includegraphics[width=\textwidth]{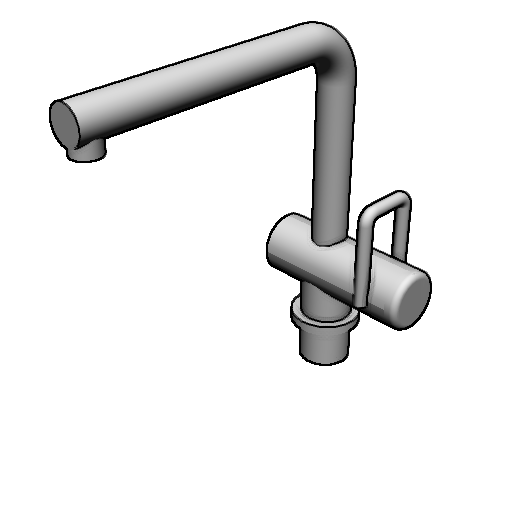}
     \end{subfigure}
     \begin{subfigure}[]{0.15\textwidth}
        \includegraphics[width=\textwidth]{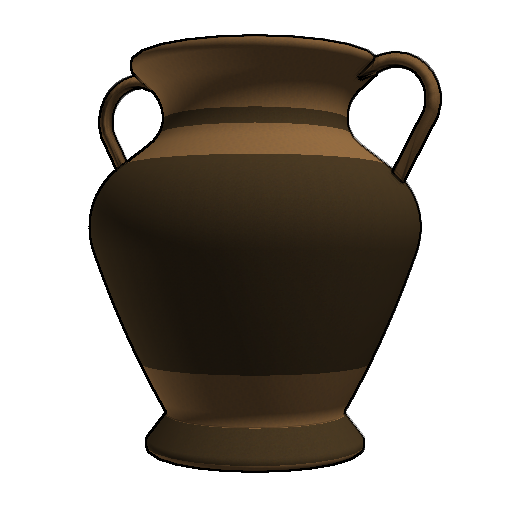}
     \end{subfigure}
     \begin{subfigure}[]{0.19\textwidth}
        \includegraphics[width=\textwidth]{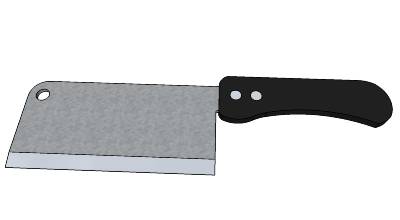}
     \end{subfigure}
     \begin{subfigure}[]{0.19\textwidth}
        \includegraphics[width=\textwidth]{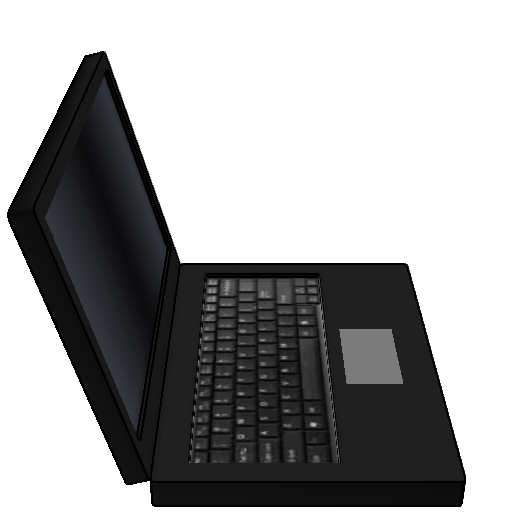}
    \end{subfigure}
     \begin{subfigure}[]{0.19\textwidth}
        \includegraphics[width=\textwidth]{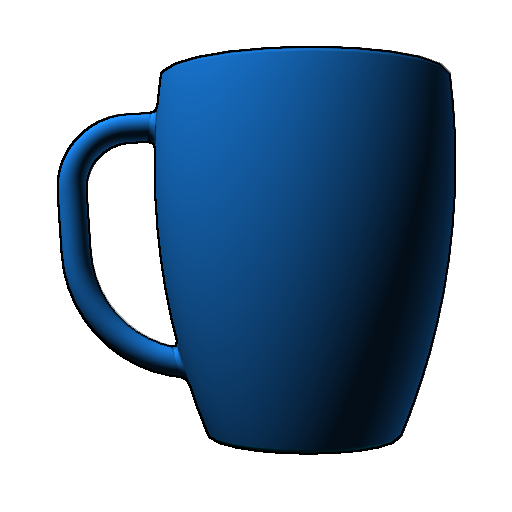}
     \end{subfigure}
     \begin{subfigure}[]{0.19\textwidth}
        \includegraphics[width=\textwidth]{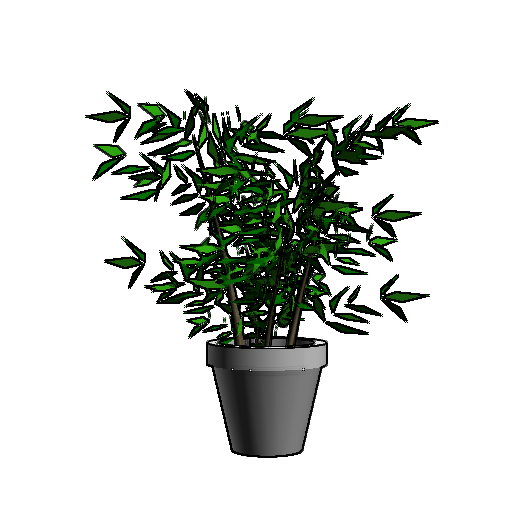}
     \end{subfigure}
     \begin{subfigure}[]{0.19\textwidth}
        \includegraphics[width=\textwidth]{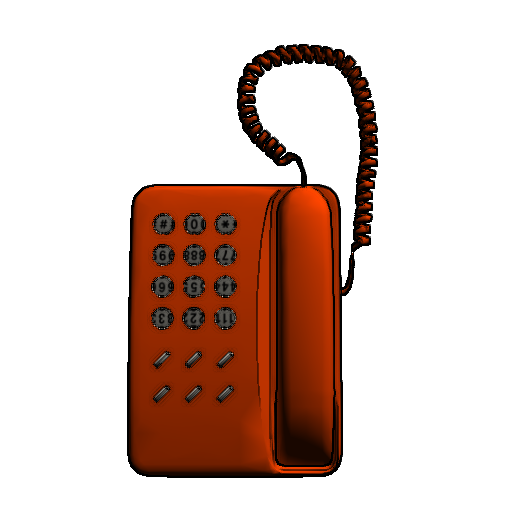}
     \end{subfigure}
     \caption{Examples of ShapeNet object assets used during the pretraining phase.}
     \label{fig:shapenet_objects}
\end{figure*}

In Figure \ref{fig:shapenet_objects} we provide images of some of the ShapeNet object assets used in the pretraining environments.

\subsection{Ego4D Textures}
\begin{figure*}[]
\centering
     \begin{subfigure}[]{0.24\textwidth}
        \includegraphics[width=\textwidth]{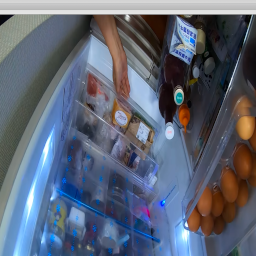}
     \end{subfigure}
     \begin{subfigure}[]{0.24\textwidth}
        \includegraphics[width=\textwidth]{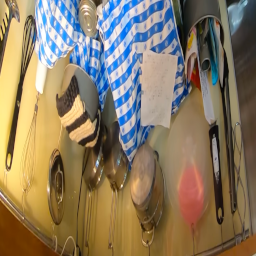}
     \end{subfigure}
     \begin{subfigure}[]{0.24\textwidth}
        \includegraphics[width=\textwidth]{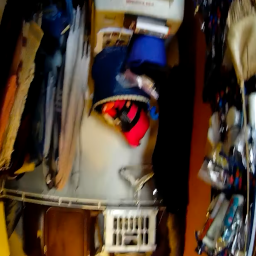}
     \end{subfigure}
     \begin{subfigure}[]{0.24\textwidth}
        \includegraphics[width=\textwidth]{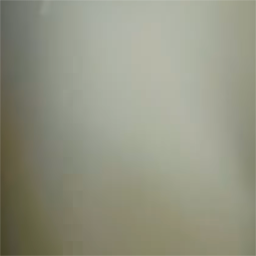}
     \end{subfigure}
     \begin{subfigure}[]{0.24\textwidth}
        \includegraphics[width=\textwidth]{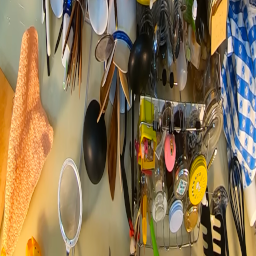}
     \end{subfigure}
     \begin{subfigure}[]{0.24\textwidth}
        \includegraphics[width=\textwidth]{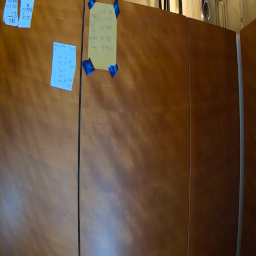}
     \end{subfigure}
     \begin{subfigure}[]{0.24\textwidth}
        \includegraphics[width=\textwidth]{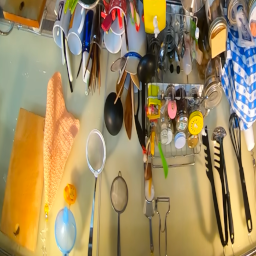}
     \end{subfigure}
     \begin{subfigure}[]{0.24\textwidth}
        \includegraphics[width=\textwidth]{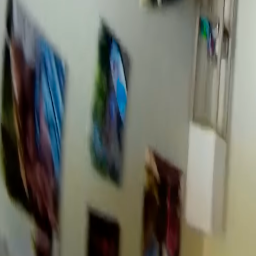}
    \end{subfigure}
    \begin{subfigure}[]{0.24\textwidth}
        \includegraphics[width=\textwidth]{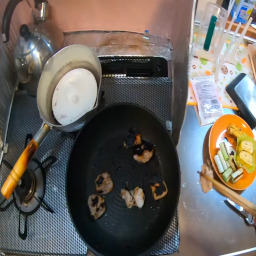}
     \end{subfigure}
     \begin{subfigure}[]{0.24\textwidth}
        \includegraphics[width=\textwidth]{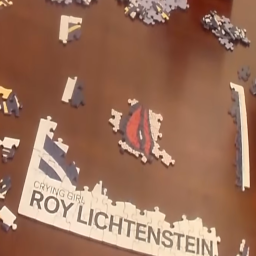}
     \end{subfigure}
     \begin{subfigure}[]{0.24\textwidth}
        \includegraphics[width=\textwidth]{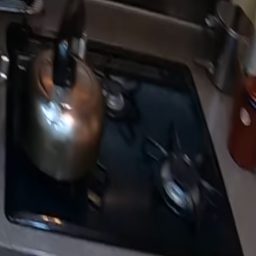}
     \end{subfigure}
     \begin{subfigure}[]{0.24\textwidth}
        \includegraphics[width=\textwidth]{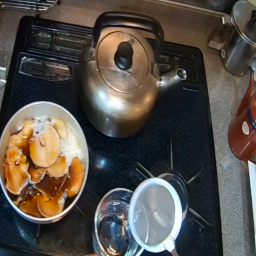}
     \end{subfigure}
     \begin{subfigure}[]{0.32\textwidth}
        \includegraphics[width=\textwidth]{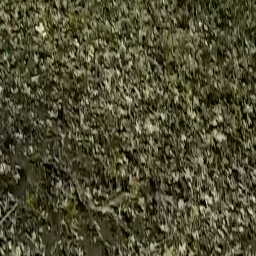}
     \end{subfigure}
     \begin{subfigure}[]{0.32\textwidth}
        \includegraphics[width=\textwidth]{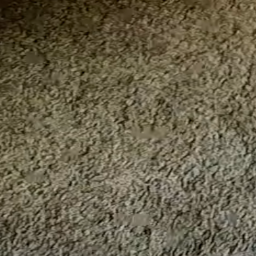}
     \end{subfigure}
     \begin{subfigure}[]{0.32\textwidth}
        \includegraphics[width=\textwidth]{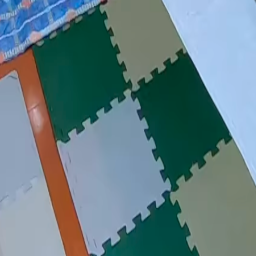}
     \end{subfigure}
     \caption{Textures cropped out of Ego4D videos and overlayed to the RLBench scene. The textures on the first two rows were overlayed to the walls, those on the third row to the table, and those on the fourth row to the floor.}
     \label{fig:ego4dtextures}
\end{figure*}

In Figure \ref{fig:ego4dtextures} we provide the textures extracted from Ego4D we overlayed on the pretraining environments.

\subsection{Compute}
We used a NVIDIA DGX A100 GPUs for all experimentation. Pretraining time for 100k steps takes 20-24 hours; finetuning for 100k takes 16-19 hours.

\section{Additional Experiments and Ablations}

\subsection{Finetuning Results with Instruction Tuning}
\begin{figure}[h!]
    \centering
    \includegraphics[width=0.32\textwidth]{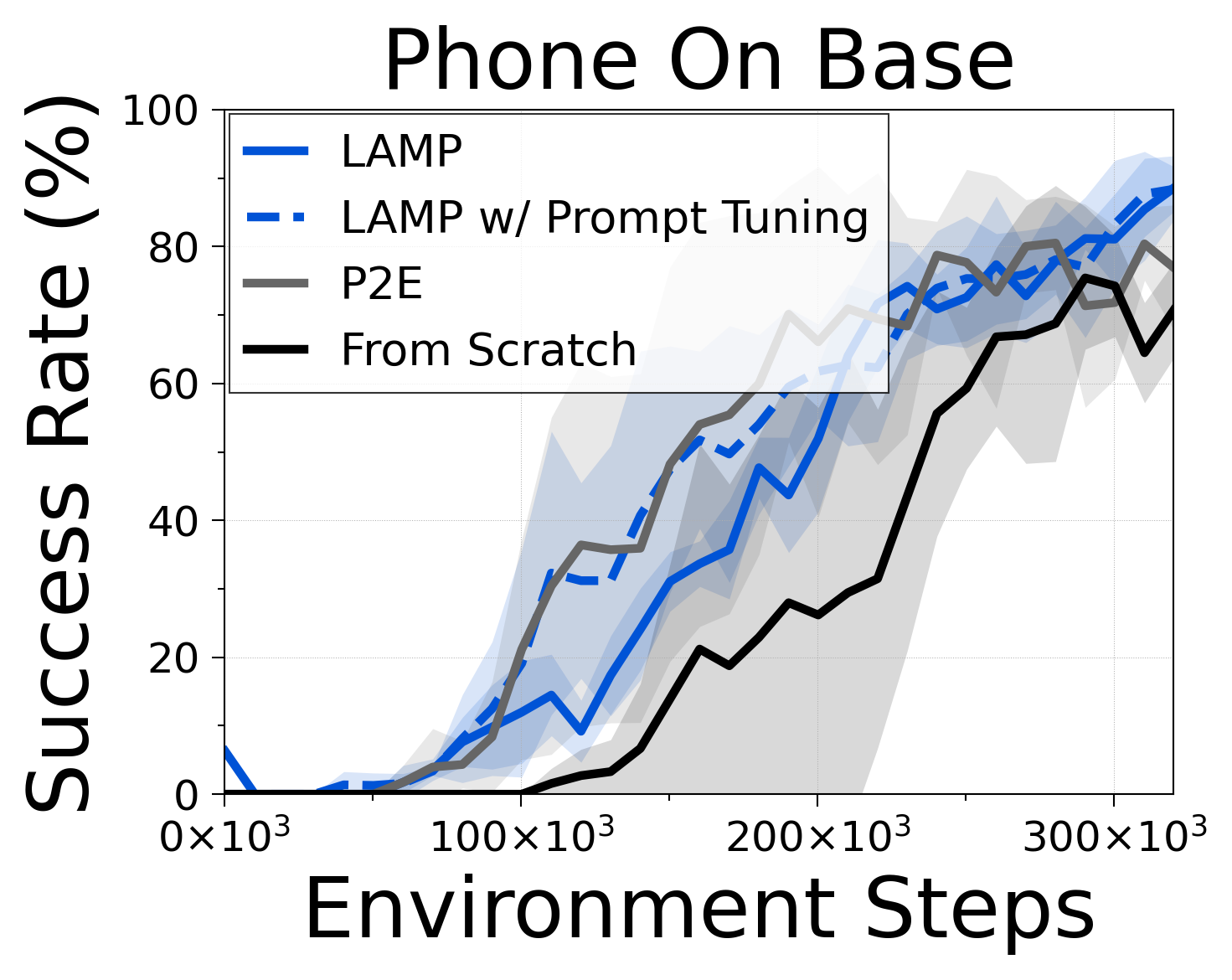}
    \caption{Finetuning Result on Phone On Base}
    \label{fig:phone_base}
\end{figure}

We include additional finetuning results for the "Phone On Base" RLBench task in Figure~\ref{fig:phone_base}. We find that by tuning the language instruction used to condition the model, we further increase performance. We provide results involving selecting a frozen language instruction from a set of many semantically similar generated instructions for finetuning. We consider 9 of these new generated prompts in addition to the original task name based prompt. We then select the language prompt corresponding to the policy that achieves the highest zero-shot evaluation return when rolled out. This simple tuning step provides us with increases in performance on the new “Phone on Base'' task. The optimal instructions tuned for each of the three seeds include:

1. The arm is picking up the phone and placing it on the base

2. The robot arm grasps the phone and sets it down

3. The robot gripper picks up the phone and places it on the base

\subsection{Reward Weighting Ablation}
\begin{figure*}[t!]
\centering
    \begin{subfigure}[]{0.32\textwidth}
        \includegraphics[width=\textwidth]{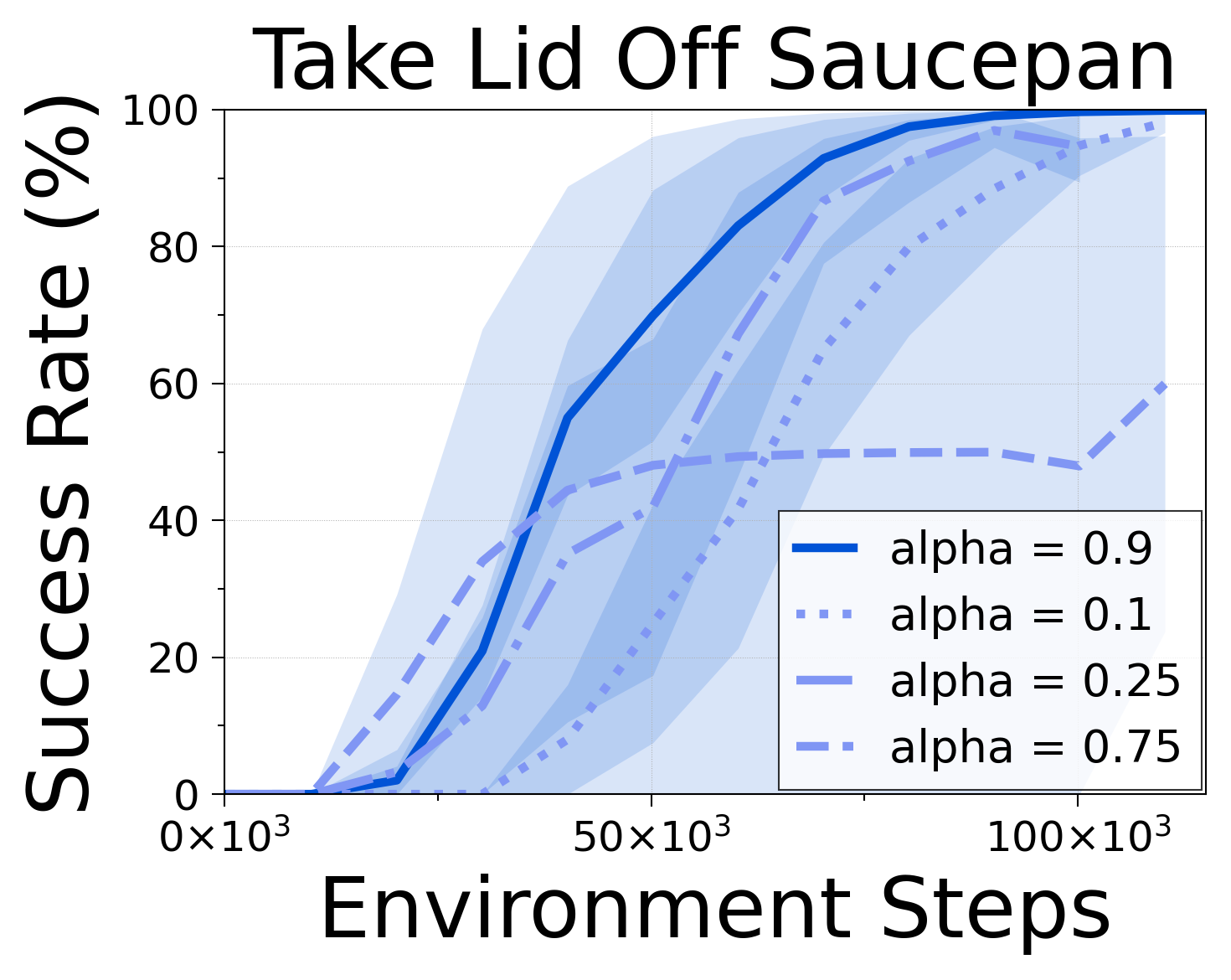}
     \end{subfigure}
     \begin{subfigure}[]{0.32\textwidth}
        \includegraphics[width=\textwidth]{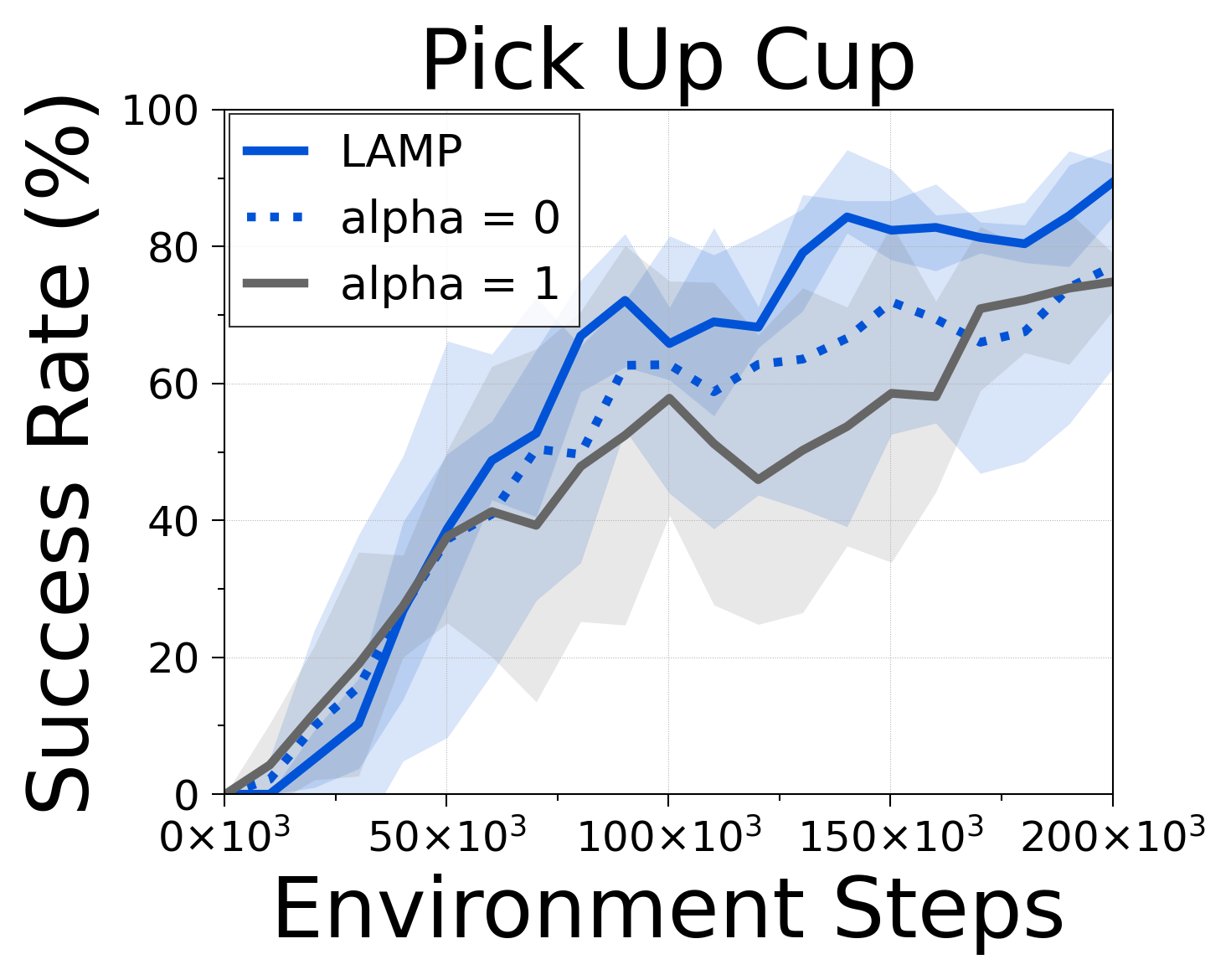}
     \end{subfigure}
     \caption{Reward Weighting Ablation}
     \label{fig:alpha}
\end{figure*}

We provide an ablation of the $\alpha$ reward weighting term for the joint pretraining reward introduced in Equation \ref{eq:lamp}. As shown in Figure \ref{fig:alpha}, we find that the method is somewhat robust to the choice of $\alpha$, however, larger values in general work better. We also isolate the relative contributions of the Plan2Explore and LAMP rewards by setting the alpha value to 0 and 1. We find that, while either option performs somewhat similarly, the synergy of the two rewards achieves the highest performance.

\subsection{Pretraining Performance}
\begin{figure}[h!]
    \centering
    \includegraphics[width=0.32\textwidth]{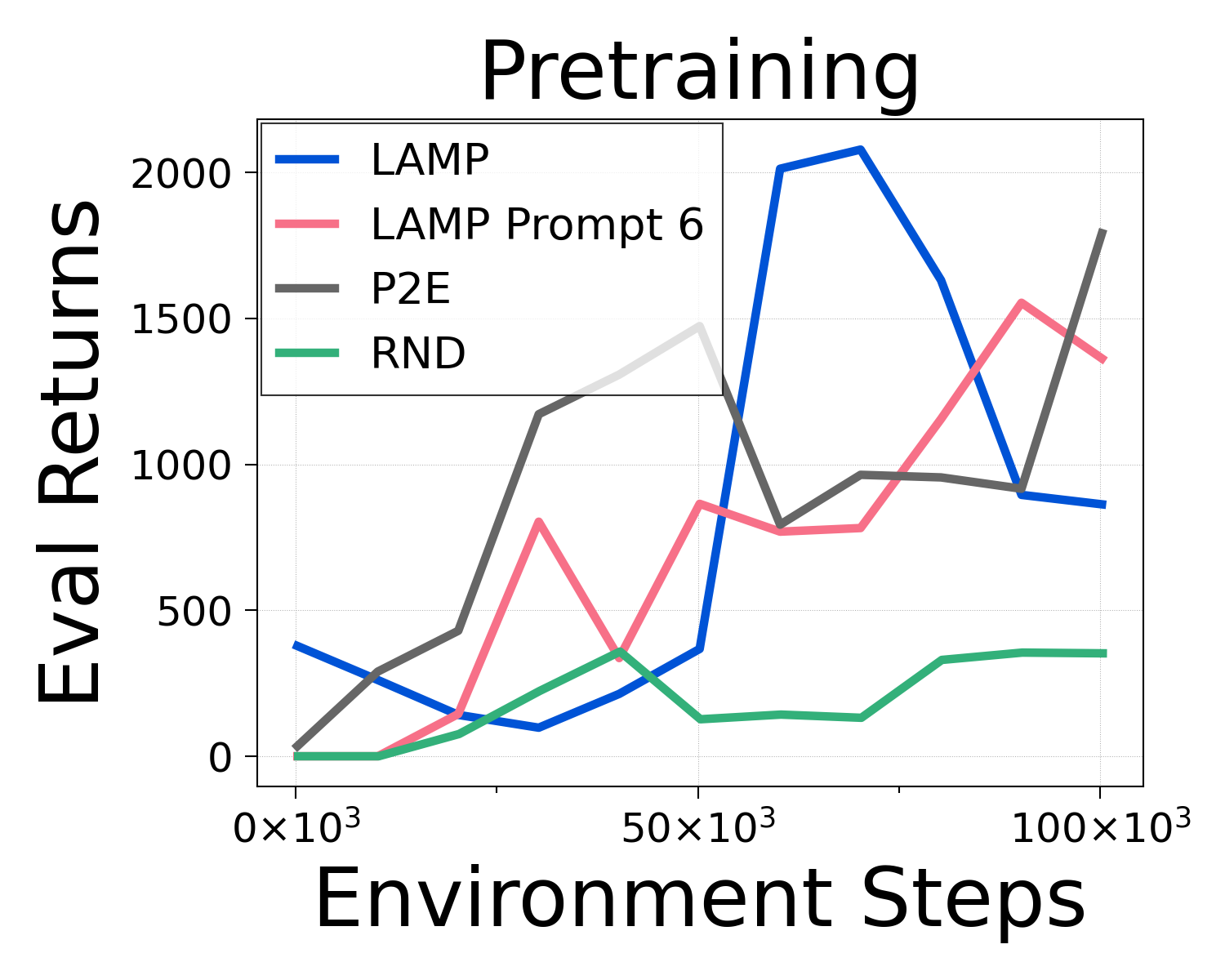}
    \caption{Pretraining Returns with different methods}
    \label{fig:pretrain}
\end{figure}

In our pretraining environment, we spawn ShapeNet objects and plot evaluation returns in Figure~\ref{fig:pretrain}. The evaluation return is based on a shaped reward for reaching the object and grasping it. While the pretraining reward is noisy, it provides insight into the types of behaviors learned, and in particular, if exploration leads to high-reward behaviors. RND is the worst performing during pretraining, and LAMP learns higher-reward behaviors through the course of training. 

\subsubsection{Random Network Distillation}
\begin{figure*}[h!]
\centering
    \begin{subfigure}[]{0.32\textwidth}
        \includegraphics[width=\textwidth]{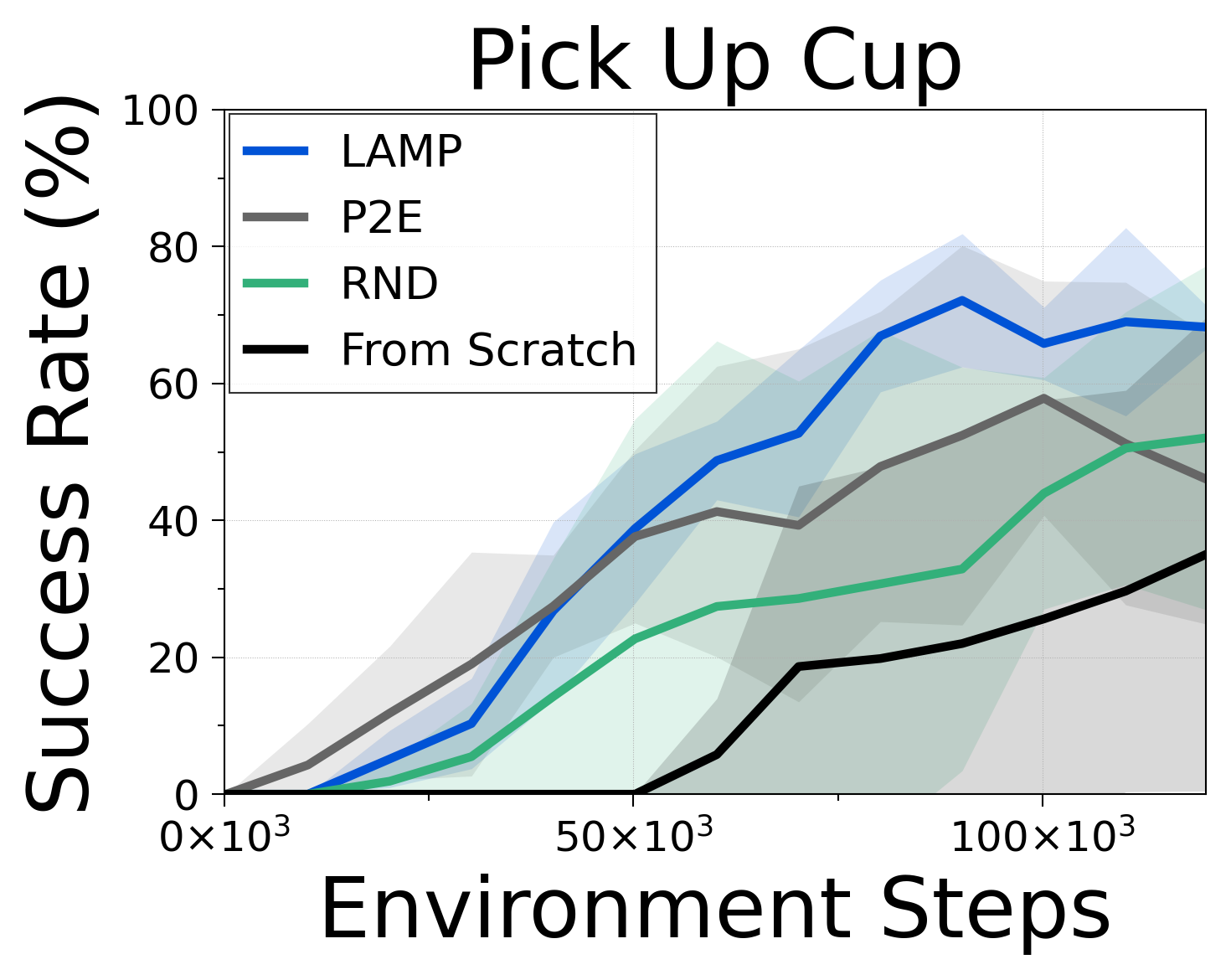}
     \end{subfigure}
     \begin{subfigure}[]{0.32\textwidth}
        \includegraphics[width=\textwidth]{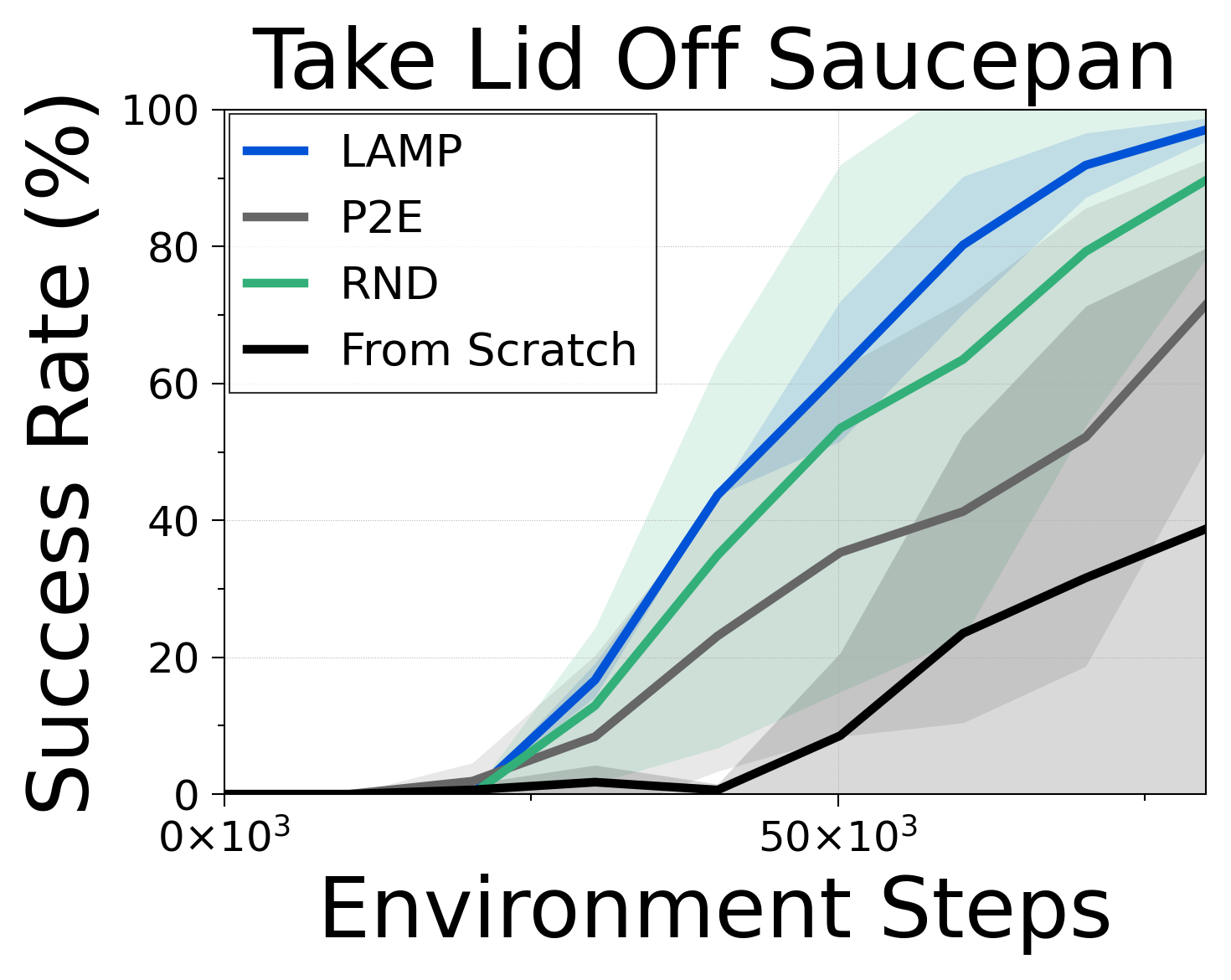}
     \end{subfigure}
     \begin{subfigure}[]{0.32\textwidth}
        \includegraphics[width=\textwidth]{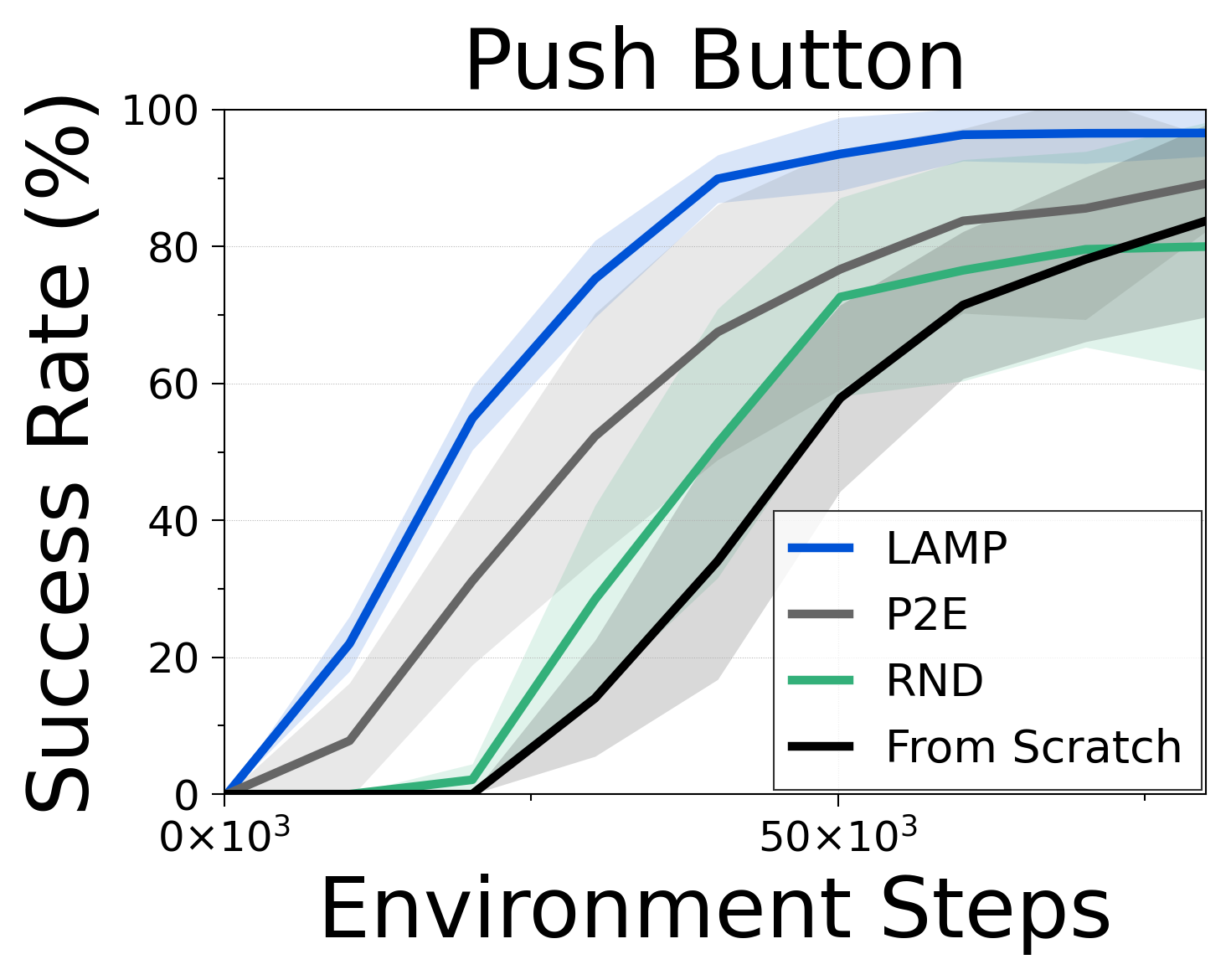}
     \end{subfigure}
     \caption{Finetuning performance on visual robotic manipulation tasks in RLBench. The solid line and shaded region represent mean and standard deviation across 3 seeds.}
     \label{fig:finetunernd}
\end{figure*}

We experiment with Random Network Distillation \citep{rnd}, an additional unsupervised reinforcement learning algorithm, as an additional baseline and report results in \ref{fig:finetunernd}. We find that for the "Pick Up Cup" and "Push Button" tasks Plan2Explore is a stronger baseline. While in the case for the "Take Lid Off Saucepan" task, RND does manage to outperform Plan2Explore, LAMP exhibits stronger performance than either baseline. We report all relevant hyperparameters in Table \ref{tab:rndtable}.

\section{Hyperparameters}
\begin{table}[htbp]
  \centering
  \caption{Plan2Explore Hyperparameters}
  \begin{tabular}{ll}
    \toprule
    \textbf{Parameter} & \textbf{Value} \\
    \midrule
    Plan2Explore & False \\
    Exploration Intrinsic Scale & 0.9 \\
    Exploration Extrinsic Scale & 0.1 \\
    Exploration Optimization & \begin{tabular}[t]{@{}l@{}}
                                Optimization: Adam \\
                                Learning Rate: 3e-4 \\
                                Epsilon: 1e-5 \\
                                Clip: 100 \\
                                Weight Decay: 1e-6 \\
                             \end{tabular} \\
    Exploration Head & \begin{tabular}[t]{@{}l@{}}
                        Layers: [512, 512, 512, 512] \\
                        Activation: ELU \\
                        Normalization: None \\
                        Distribution: MSE \\
                      \end{tabular} \\
    Exploration Reward Normalization & \begin{tabular}[t]{@{}l@{}}
                                        Momentum: 1.0 \\
                                        Scale: 1.0 \\
                                        Epsilon: 1e-8 \\
                                      \end{tabular} \\
    Disaggregation Target & Stochastic \\
    Disaggregation Log & False \\
    Disaggregation Models & 10 \\
    Disaggregation Offset & 1 \\
    Disaggregation Action Condition & True \\
    Exploration Model Loss & KL \\
    \bottomrule
  \end{tabular}
  \label{tab:p2etable}
\end{table}

\begin{table}[htbp]
  \centering
  \caption{PTMae Hyperparameters}
  \begin{tabular}{ll}
    \toprule
    \textbf{Parameter} & \textbf{Value} \\
    \midrule
    MAE Image Width Size & 224 \\
    MAE Image Height Size & 224 \\
    WM Flat VIT Image Height Size & 7 \\
    WM Flat VIT Image Width Size & 7 \\
    MAE State Prediction & False \\
    WM Flat VIT Input Channels & 768 \\
    WM Flat VIT Embedding Dimension & 128 \\
    MAE Average & True \\
    \bottomrule
  \end{tabular}
  \label{tab:ptmaetable}
\end{table}

\begin{table}[htbp]
  \centering
  \caption{MAE Hyperparameters}
  \begin{tabular}{ll}
    \toprule
    \textbf{Parameter} & \textbf{Value} \\
    \midrule
    Camera Keys & `image front|image wrist' \\
    Mask Ratio & 0.95 \\
    MAE & \begin{tabular}[t]{@{}l@{}}
            Image Height Size: 128 \\
            Image Width Size: 128 \\
            Patch Size: 16 \\
            Embedding Dimension: 256 \\
            Depth: 8 \\
            Number of Heads: 4 \\
            Decoder Embedding Dimension: 256 \\
            Decoder Depth: 6 \\
            Decoder Number of Heads: 4 \\
            Reward Prediction: True \\
            Early Convolution: True \\
            State Prediction: True \\
            Input Channels: 3 \\
            Number of Cameras: 0 \\
            State Dimension: 10 \\
            View Masking: True \\
            Control Input: `front wrist' \\
          \end{tabular} \\
    WM Flat VIT & \begin{tabular}[t]{@{}l@{}}
                      Image Height Size: 8 \\
                      Image Width Size: 8 \\
                      Patch Size: 1 \\
                      Embedding Dimension: 128 \\
                      Depth: 2 \\
                      Number of Heads: 4 \\
                      Decoder Embedding Dimension: 128 \\
                      Decoder Depth: 2 \\
                      Decoder Number of Heads: 4 \\
                      Input Channels: 256 \\
                      State Prediction: False \\
                   \end{tabular} \\
    Image Time Size & 4 \\
    Use ImageNet MAE & False \\
    MAE Chunk & 1 \\
    MAE Average & False \\
    \bottomrule
  \end{tabular}
  \label{tab:maetable}
\end{table}

\begin{table}[htbp]
  \centering
  \caption{World Model Hyperparameters}
  \begin{tabular}{ll}
    \toprule
    \textbf{Parameter} & \textbf{Value} \\
    \midrule
    Grad Heads & [Reward, Discount] \\
    Predictive Discount & True \\
    RSSM & \begin{tabular}[t]{@{}l@{}}
              Action Free: False \\
              Hidden: 1024 \\
              Deterministic: 1024 \\
              Stochastic: 32 \\
              Discrete: 32 \\
              Activation: ELU \\
              Normalization: None \\
              Stochastic Activation: Sigmoid2 \\
              Minimum Standard Deviation: 0.1 \\
           \end{tabular} \\
    Reward Head & \begin{tabular}[t]{@{}l@{}}
                     Layers: [512, 512, 512, 512] \\
                     Activation: ELU \\
                     Normalization: None \\
                     Distribution: Symlog \\
                  \end{tabular} \\
    Discount Head & \begin{tabular}[t]{@{}l@{}}
                        Layers: [512, 512, 512, 512] \\
                        Activation: ELU \\
                        Normalization: None \\
                        Distribution: Binary \\
                     \end{tabular} \\
    Loss Scales & \begin{tabular}[t]{@{}l@{}}
                      Feature: 1.0 \\
                      KL: 1.0 \\
                      Reward: 1.0 \\
                      Discount: 1.0 \\
                      Proprio: 1.0 \\
                      MAE Reward: 1.0 \\
                   \end{tabular} \\
    KL & \begin{tabular}[t]{@{}l@{}}
              Scale: 1.0 \\
           \end{tabular} \\
    KL Minloss & 0.1 \\
    KL Balance & 0.8 \\
    Model Optimization & \begin{tabular}[t]{@{}l@{}}
                            Optimization: Adam \\
                            Learning Rate: 3e-4 \\
                            Epsilon: 1e-5 \\
                            Clip: 100.0 \\
                            Weight Decay: 1e-6 \\
                            Weight Decay Pattern: `kernel' \\
                            Warmup: 0 \\
                         \end{tabular} \\
    MAE Optimization & \begin{tabular}[t]{@{}l@{}}
                          Optimization: Adam \\
                          Learning Rate: 3e-4 \\
                          Epsilon: 1e-5 \\
                          Clip: 100.0 \\
                          Weight Decay: 1e-6 \\
                          Warmup: 2500 \\
                       \end{tabular} \\
    \bottomrule
  \end{tabular}
  \label{tab:wmtable}
\end{table}

\begin{table}[htbp]
  \centering
  \caption{Actor Critic Hyperparameters}
  \begin{tabular}{ll}
    \toprule
    \textbf{Parameter} & \textbf{Value} \\
    \midrule
    Actor & \begin{tabular}[t]{@{}l@{}}
              Layers: [512, 512, 512, 512] \\
              Activation: ELU \\
              Normalization: None \\
              Distribution: Trunc\_Normal \\
              Minimum Standard Deviation: 0.1 \\
           \end{tabular} \\
    Critic & \begin{tabular}[t]{@{}l@{}}
               Layers: [512, 512, 512, 512] \\
               Activation: ELU \\
               Normalization: None \\
               Distribution: MSE \\
           \end{tabular} \\
    Actor Optimization & \begin{tabular}[t]{@{}l@{}}
                           Optimization: Adam \\
                           Learning Rate: 1e-4 \\
                           Epsilon: 1e-5 \\
                           Clip: 100.0 \\
                           Weight Decay: 1e-6 \\
                           Weight Decay Pattern: `kernel' \\
                           Warmup: 0 \\
                        \end{tabular} \\
    Critic Optimization & \begin{tabular}[t]{@{}l@{}}
                            Optimization: Adam \\
                            Learning Rate: 1e-4 \\
                            Epsilon: 1e-5 \\
                            Clip: 100.0 \\
                            Weight Decay: 1e-6 \\
                            Weight Decay Pattern: `kernel' \\
                            Warmup: 0 \\
                         \end{tabular} \\
    Discount & 0.99 \\
    Discount Lambda & 0.95 \\
    Image Horizon & 15 \\
    Actor Grad & Dynamics \\
    Actor Grad Mix & 0.1 \\
    Actor Entropy & \begin{tabular}[t]{@{}l@{}}
              Scale: 1e-4 \\
           \end{tabular} \\
    Slow Target & True \\
    Slow Target Update & 100 \\
    Slow Target Fraction & 1 \\
    Slow Baseline & True \\
    Reward Normalization & \begin{tabular}[t]{@{}l@{}}
                              Momentum: 0.99 \\
                              Scale: 1.0 \\
                              Epsilon: 1e-8 \\
                            \end{tabular} \\
    \bottomrule
  \end{tabular}
  \label{tab:actorcritictable}
\end{table}

\begin{table}[htbp]
  \centering
  \caption{Random Network Distillation Hyperparameters}
  \begin{tabular}{ll}
    \toprule
    \textbf{Parameter} & \textbf{Value} \\
    \midrule
    Embedding Dimension & 512 \\
    Hidden Dimension & 256 \\
    Optimizer & \begin{tabular}[t]{@{}l@{}}
                     Adam \\
                     Learning Rate: 3e-4 \\
                     Epsilon: 1e-5 \\
                     Clip: 100.0 \\
                     Weight Decay: 1e-6 \\
                     Warmup: 2500 \\
               \end{tabular} \\
    \bottomrule
  \end{tabular}
  \label{tab:rndtable}
\end{table}

\end{document}